\documentclass{article} 
\usepackage{iclr2026_conference,times}

\usepackage{amsmath,amsfonts,bm}

\def\eqref#1{equation~\ref{#1}}

\def\1{\bm{1}}

\DeclareMathAlphabet{\mathsfit}{\encodingdefault}{\sfdefault}{m}{sl}
\SetMathAlphabet{\mathsfit}{bold}{\encodingdefault}{\sfdefault}{bx}{n}

\usepackage{hyperref}
\usepackage{url}

\usepackage{amsmath}

\usepackage{graphicx}
\usepackage{tikz}
\usetikzlibrary{positioning}
\usepackage{cleveref}

\usepackage{makecell} %
\usepackage{multirow}
\usepackage{booktabs} %

\def \eg {\emph{e.g.}}

\title{%
    \includegraphics[width=0.95em]{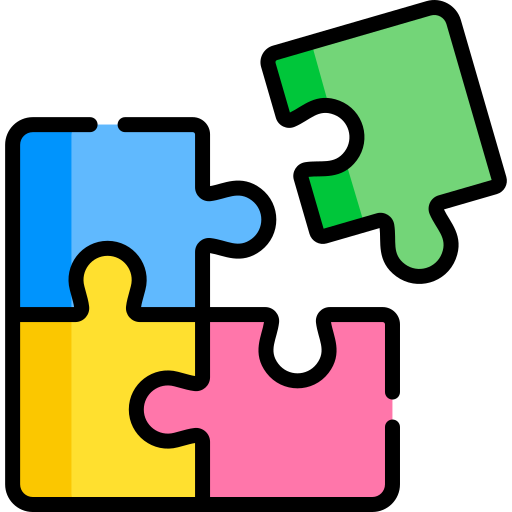}~%
    Jigsaw3D: Disentangled 3D Style Transfer via Patch Shuffling and Masking
}

\iclrfinalcopy

\author{
    {\bf Yuteng Ye\textsuperscript{1}},
    {\bf Zheng Zhang\textsuperscript{1}},
    {\bf Qinchuan Zhang\textsuperscript{1}},
    {\bf Di Wang\textsuperscript{1}},
    {\bf Youjia Zhang\textsuperscript{2}}, 
    {\bf Wenxiao Zhang\textsuperscript{1}}, \\ 
    {\bf Wei Yang\textsuperscript{2}},
    {\bf Yuan Liu\textsuperscript{3}} \\
    \\
    \textsuperscript{1}Huawei, Media Innovation Lab \\
    \textsuperscript{2}Huazhong University of Science \& Technology \\
    \textsuperscript{3}The Hong Kong University of Science and Technology
}

\begin{document}
\maketitle

\begin{figure*}[h] 
    \centering
    \vspace{-0.5em} %
    \includegraphics[width=0.98\textwidth]{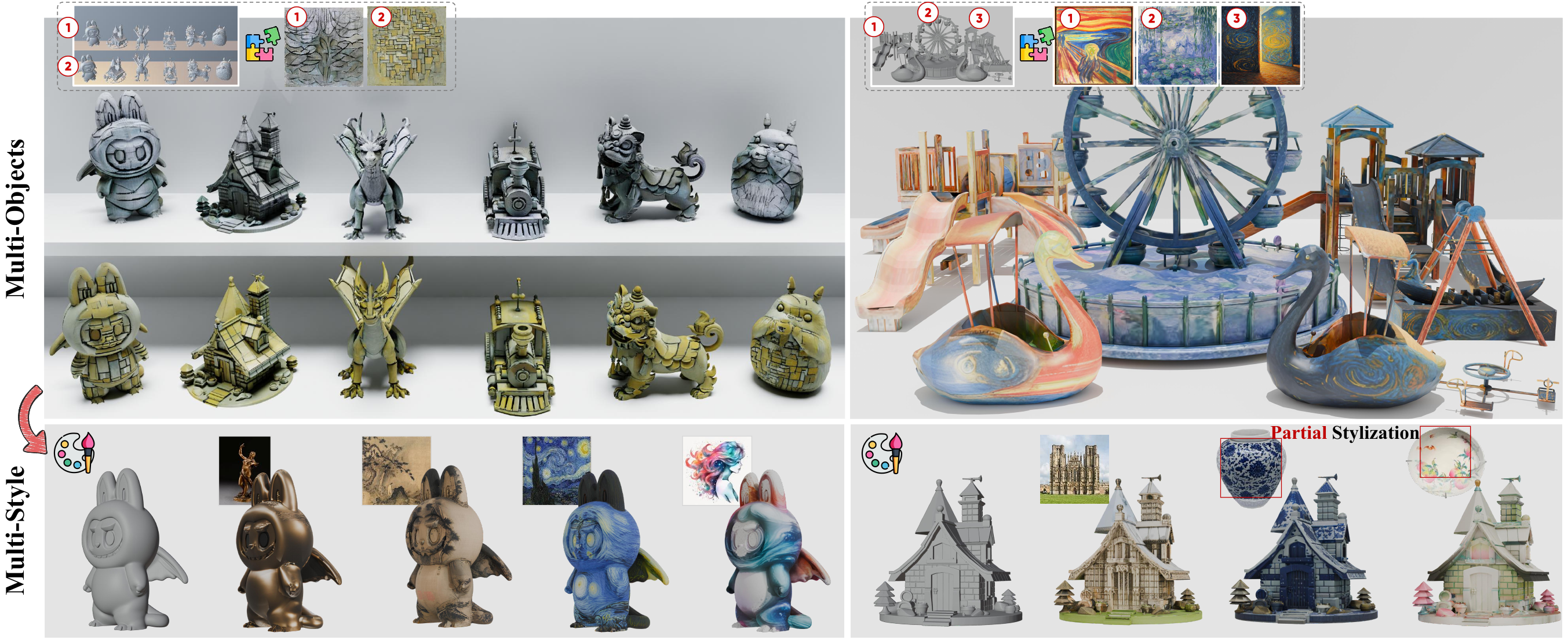}
    \caption{
        We propose the JIGSAW3D, a versatile 3D stylization framework that transfers stylistic statistics from 2D images to 3D meshes. Our method achieves high stylistic consistency across multiple, diverse objects in a scene (\textbf{top}). Furthermore, it demonstrates high versatility with various art styles and supports partial reference stylization for fine-grained user control (\textbf{bottom}). 
    }
    \label{fig:teaser}
    \vspace{-0.5em} %
\end{figure*}

\begin{abstract}
Controllable 3D style transfer seeks to restyle a 3D asset so that its textures match a reference image while preserving the integrity and multi-view consistency. The pravelent methods either rely on direct reference style token injection or score-distillation from 2D diffusion models, which incurs heavy per-scene optimization and often entangles style with semantic content. We introduce Jigsaw3D, a multi-view diffusion based pipeline that decouples style from content and enables fast, view-consistent stylization. Our key idea is to leverage the jigsaw operation—spatial shuffling and random masking of reference patches—to suppress object semantics and isolate stylistic statistics (color palettes, strokes, textures). We integrate these style cues into a multi-view diffusion model via reference-to-view cross-attention, producing view-consistent stylized renderings conditioned on the input mesh. The renders are then style-baked onto the surface to yield seamless textures. Across standard 3D stylization benchmarks, Jigsaw3D achieves high style fidelity and multi-view consistency with substantially lower latency, and generalizes to masked partial reference stylization, multi-object scene styling, and tileable texture generation. Project page is available at: https://babahui.github.io/jigsaw3D.github.io/
\end{abstract}

\section{Introduction}

The field of 3D object generation has advanced rapidly, driven by progress in 3D generative modeling~\citep{zhang2024clay,xiang2025structured,li2025sparc3d,wu2024direct3d,zhang20233dshape2vecset}, neural representations~\citep{mildenhall2021nerf,park2019deepsdf,mescheder2019occupancy}, and the availability of large-scale 3D datasets~\citep{chang2015shapenet,deitke2023objaverse}. Within this landscape, 3D stylization—transferring the artistic style of a 2D reference image to a 3D asset while preserving object identity and multi-view consistency—has emerged as a practical requirement for virtual reality, game development, and animated content creation. Progress, however, is hampered by the absence of large-scale 3D style corpora containing paired texture and style supervision, which makes end-to-end supervised training impractical.

In response, recent approaches typically follow one of two directions. Training-free methods inject reference style cues into frozen attention layers to achieve multi-view style fusion; for example, 3D-style-LRM~\citep{oztas20253d} combines CLIP-derived features within attention modules, and Style3D~\citep{song2024style3d} modifies self-attention keys/values to propagate style across views. Score-distillation strategies instead leverage diffusion-based style objectives to fine-tune neural rendering pipelines (e.g., StyleTex~\citep{xie2024styletex}). While effective in limited settings, these paradigms often (i) struggle to disentangle style from semantic content—leading to texture leakage and degraded geometry/appearance fidelity—or (ii) require computationally expensive, per-asset optimization, limiting scalability.

To address the lack of explicit ``style–texture'' image pairs for training 3D stylization models, we revisit what constitutes a style reference. Conventional practice treats a ``reference style image'' as a natural image that entangles global semantics (object layout, parts, viewpoint) with style attributes (color palette, stroke-like texture, frequency statistics), which in turn requires substantial variation in both content and style for effective learning. We instead posit that an effective style reference should convey style independently of semantics, and that local image patches are sufficient carriers of such statistics~\citep{wang2023styleadapter}.
Building on this insight, we introduce a jigsaw transform that randomly shuffles and sparsely masks non-overlapping patches, destroying global structure while preserving local style cues. This enables us to synthesize style–texture supervision from textured 3D assets. Concretely, given a textured asset (e.g., from Objaverse~\citep{deitke2023objaverse}), we render multi-view images, apply the jigsaw transform to one rendered view to obtain a semantics-agnostic style reference, and use the remaining views as supervision targets. This procedure yields large quantities of pseudo-paired data without requiring curated style–texture pairs.
We then train a multi-view stylized image generator conditioned on the style reference and geometry. The style image is encoded by a pretrained text-to-image (T2I) diffusion network; intermediate activations are extracted as disentangled style conditions. These conditions guide a multi-view diffusion model built on a U-Net backbone that integrates three complementary attention mechanisms: (a) self-attention for intra-view coherence, (b) multi-view attention to enforce cross-view consistency, and (c) reference attention that injects the style conditions to perform dynamic, style-adaptive feature recombination. In addition, geometric signals (normal and position maps) are processed by a conditional encoder and injected into the U-Net’s spatial features to respect object geometry during generation. By construction, the jigsawed reference suppresses content leakage while retaining style statistics, enabling explicit content–style disentanglement and scalable training without per-asset optimization or manually paired 3D style datasets.
Finally, we style-bake the multi-view outputs into a textured asset: reproject stylized views to the UV atlas with visibility/z-tests and fuse per-texel observations via seam-aware, confidence-weighted blending to obtain albedo. We then bake tangent-space normals and complete missing texels with UV-space inpainting, yielding a complete, cross-view-consistent texture.

Experiments demonstrate that our method attains state-of-the-art results on multiple 3D stylization benchmarks and generalizes to partial stylization, multi-object scene styling, and tileable texture generation, without per-asset optimization.

The main contributions of this work are summarized as follows:
\begin{itemize}
\item \textbf{Jigsaw-based style reference construction.} We introduce a semantics-destroying jigsaw transform—spatial patch shuffling with random masking—that disentangles style from content and synthesizes style–texture pseudo-pairs from textured 3D assets for supervised stylization training.
\item \textbf{Reference-attention for stylization.} We design a trainable reference-attention module that injects disentangled style conditions to enable dynamic, style-aware feature recombination.
\end{itemize}

\section{Related Work}
\textbf{3D Texture Generation.}
3D texture generation aims to create visually consistent and semantically meaningful texture maps for 3D objects. A central challenge in this task lies in achieving multi-view geometric consistency and maintaining texture coherence across different viewpoints. Some approaches relied on optimization-based methods~\citep{poole2022dreamfusion,lin2023magic3d} that leverage pre-trained 2D diffusion models~\citep{rombach2022high} through score distillation sampling, suffering high computational costs.
More related to our work are methods focusing on novel view generation with T2I models. These approaches build upon text-to-image diffusion models~\citep{rombach2022high}, which provide a strong prior of 2D image appearance, and extend them to generate geometrically consistent multi-view images. Zero-1-to-3~\citep{liu2023zero} serves as a foundational model that predicts novel views from a single image using viewpoint-conditioned diffusion. MVDream~\citep{shi2023MVDream} extends this further by injecting camera parameters into the self-attention mechanism, enabling explicit 3D awareness and cross-view consistency. SyncDreamer~\citep{liu2023syncdreamer} introduces synchronized multi-view generation through feature-level fusion across views, while MV-Adapter~\citep{huang2024mv} employs lightweight adapter modules to efficiently fine-tune pre-trained T2I models for multi-view synthesis. These methods commonly integrate camera embeddings or geometric constraints to maintain multi-view consistency while leveraging large-scale 3D data~\citep{deitke2023objaverse,deitke2023objaverse-xl} for training.

\textbf{Image-Guided Stylization.}
Image-guided stylization aims to transfer the style from a reference image to a target while preserving its semantic structure. A central challenge lies in effectively representing and transferring style features. Early approaches typically relied on statistical summarization of deep features, such as Gram matrices~\citep{gatys2016image} or channel-wise mean and variance alignment~\citep{huang2017arbitrary,lu2019closed}. With recent advancements, research in this area has evolved along two main directions: 2D and 3D style transfer approaches.

In the domain of 2D style transfer, the rise of diffusion models has spurred the development of various fine-tuning strategies. These include full model fine-tuning~\citep{zhang2023adding,zhang2022domain}, lightweight adapter-based approaches~\citep{wang2023styleadapter,mou2024t2i,ye2023ip} that insert trainable modules into pre-trained networks, and low-rank adaptation (LoRA)~\citep{hu2022lora,frenkel2024implicit} that captures style characteristics via weight updates. 
More recently, attention-based methods have attracted increasing interest. Among these, StyleAligned~\citep{hertz2024style} ensures consistent style across generated images by sharing self-attention and aligning the query and key features of target images with a reference via AdaIN~\citep{huang2017arbitrary}. Visual Style Prompting~\citep{jeong2024visual} enables training-free style transfer by replacing the key and value features in the target's self-attention.
Building on these attention designs, StyleAdapter~\citep{wang2023styleadapter} reduces semantic interference by removing feature class tokens and shuffling positional embeddings. Although the shuffling strategy in~\citep{wang2023styleadapter,gu2018arbitrary} indicates that style information can be preserved within feature patches, it has not been applied to 3D stylization tasks.

Research progress in 3D-aware stylization remains considerably limited compared to 2D stylization.
Previous NeRF-based approaches~\citep{fujiwara2024style} typically depend on multi-view images and require per-asset optimization. StyleTex~\citep{xie2024styletex} decomposes style diffusion loss via orthogonal projection in a semantic-aware feature space, yet its test-time rendering optimization incurs significant computational overhead.
Other training-free methods inject style features through attention fusion. Style3D~\citep{song2024style3d} directly transfers self-attention features from a 2D reference image to multi-view generation. 3D-style-LRM~\citep{oztas20253d} integrates style information through linear combinations of CLIP-based reference features.

In contrast to existing methods, we introduce a jigsaw-based disentanglement strategy to create style-texture pairs, enabling the training of dynamic style-aware feature recombination. To the best of our knowledge, our work presents the first approach to incorporate image-jigsaw for 3D stylization.

\section{Methods}
\begin{figure}[h]
	\centering
        \includegraphics[width=0.98\textwidth]{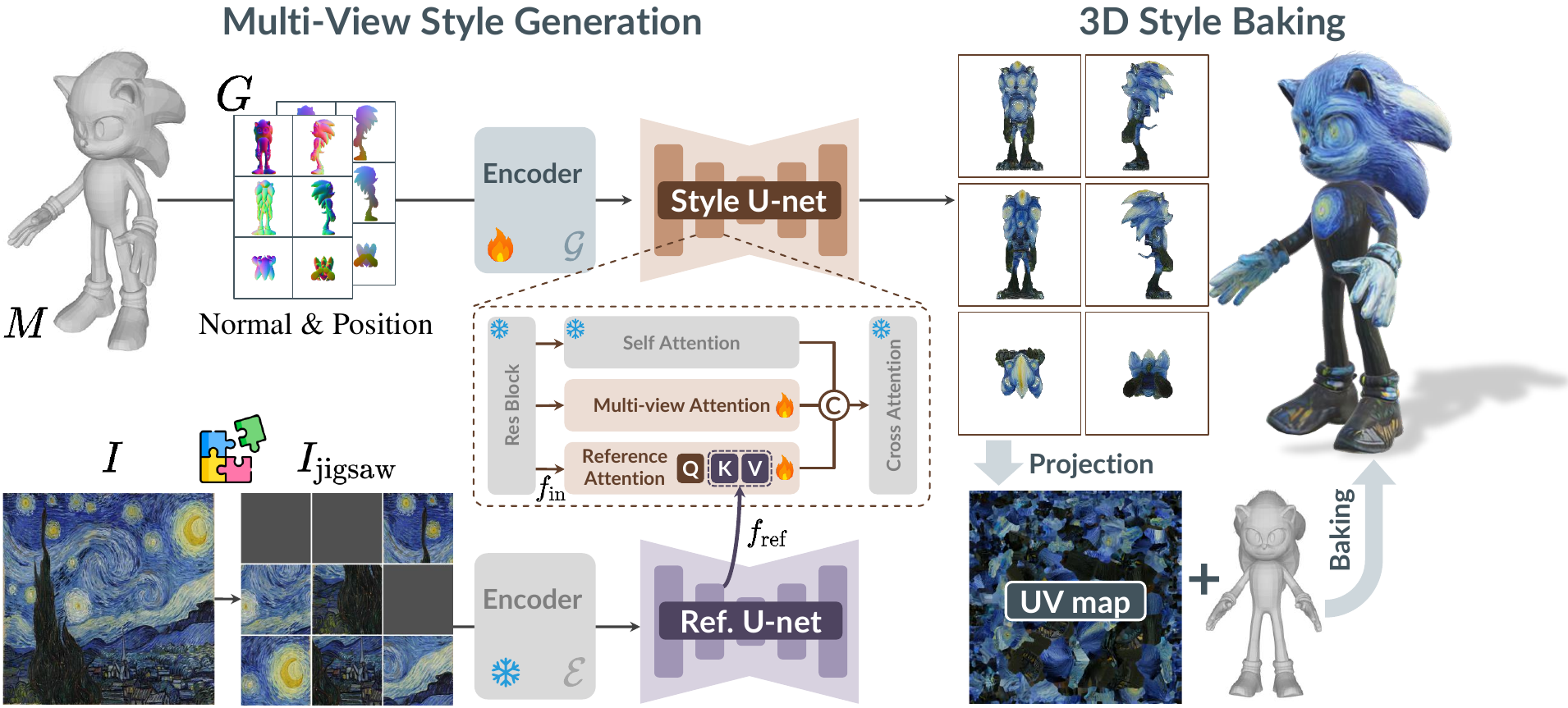} 
	\caption{\textbf{Our Method Pipeline.} The whole framework contains multi-view stylized image generation and 3D style baking. \textbf{Multi-View Style Generation:} position and normal maps from the mesh \(M\) are encoded and injected into a style U-Net via feature modulation, while the reference image \(I\) is processed by a jigsaw operation involving image patch shuffling and random masking to extract style features. These style features are sent to a pre-trained reference U-Net to extract intermediate features that serve as keys and values in a reference attention module. Our style U-Net uses reference attention for aligning with the reference style and multi-view attention to ensure cross-view consistency. \textbf{3D Style Baking:} The generated multi-view images are projected onto the mesh's UV space, yielding a seamless UV map ready for final rendering.}
    \label{fig:pipeline}
\end{figure}

Our method first constructs style-texture pairs from existing 3D assets to serve as training data for the framework (Sec.~\ref{sec:dataset}). Subsequently, it operates to generate multi-view stylized images and bakes these views into a stylized 3D object (Sec.~\ref{sec:pipeline}, see~\Cref{fig:pipeline}).

\subsection{Style-Texture Pairs Creation}
\label{sec:dataset}
Unlike heavy score distillation-based approaches~\citep{fujiwara2024style,song2024style3d}, our method adopts a data-driven manner by constructing a style-3D dataset, enabling the model to acquire stylization transfer capabilities through supervised training.
However, current large-scale 3D datasets such as Objaverse~\citep{deitke2023objaverse} typically exhibit complex representations where semantic content and style attributes are intricately entangled within texture maps, making style extraction particularly challenging.
A critical initial step involves developing an effective approach to disentangle style information from texture maps.

\begin{figure}[h]
\centering
\begin{minipage}{0.45\textwidth}
\centering
\includegraphics[width=\textwidth]{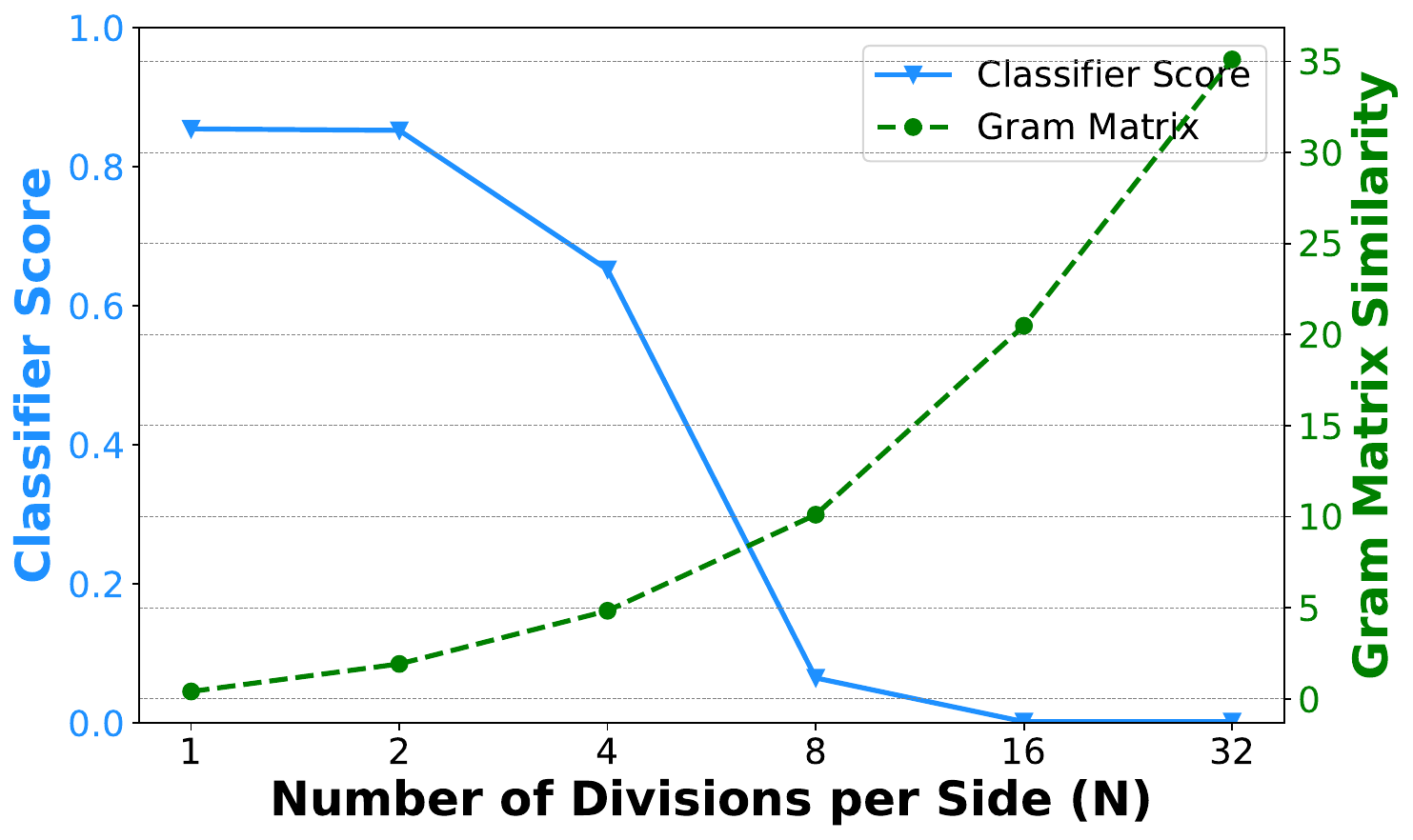}
\end{minipage}
\hfill
\begin{minipage}{0.54\textwidth}
\centering
\includegraphics[width=\textwidth]{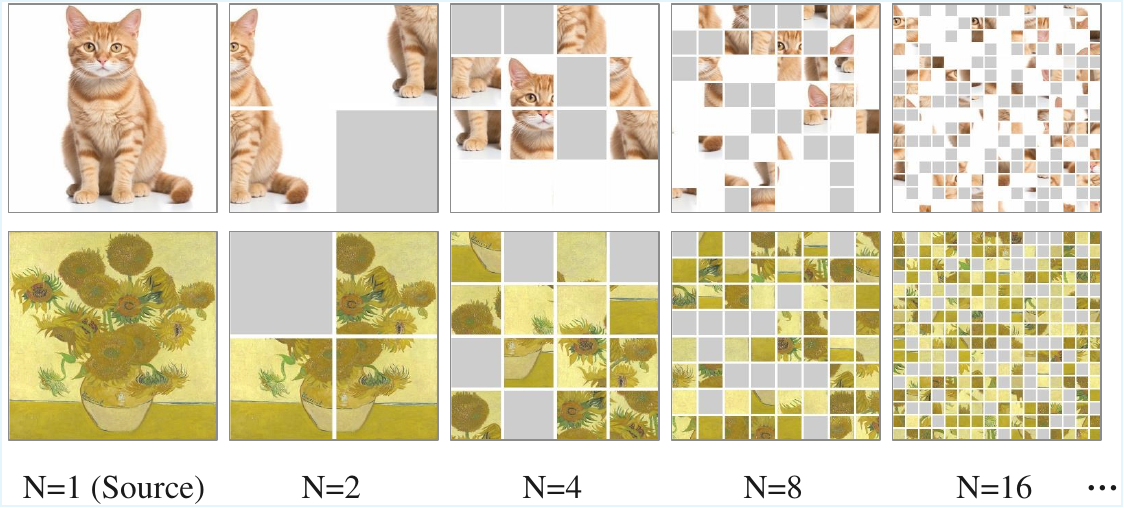}
\end{minipage}
\caption{
Analysis of style-content disentanglement through patch shuffling and masking. We apply different degrees of shuffling and a fixed mask ratio.
\textbf{Left:} Quantitative evaluation of content and style attributes under increasing shuffle intensity. As $N$ (number of divisions per image side) increases, the CNN-based classification score (blue line) of shuffled images decreases sharply. At $N=8$, semantic content is almost entirely lost. Meanwhile, the Gram matrix similarity~\citep{gatys2016image} (denoted as green dashed line) calculated between shuffled images and source images increases gradually for $N \leq 8$, indicating well-preserved style fidelity. The setting $N=8$ strikes a good balance between semantic suppression and style preservation.
\textbf{Right:} Visual examples of shuffled images using different values of $N$ and a fixed mask ratio.
}
\label{fig:motivation}
\end{figure}

\textbf{Disentanglement Motivation.}
Generally disentanglement requires finding a common representation to express variations among different statistical dimensions. It has been observed in 2D style transfer that image patches can serve as effective carriers of style information~\citep{wang2023styleadapter}. Based on this insight, we further posit that deliberately shuffling and masking image patches can disrupt object structures and suppress global semantics. At the same time, such patch-level shuffling preserves first- and second-order style statistics (\eg~mean and variance)~\citep{huang2017arbitrary}. This behavior is further quantitatively demonstrated in \Cref{fig:motivation}, where beyond a certain shuffling intensity, semantic content is largely eliminated while style fidelity remains well-preserved.
Motivated by these observations, we introduce a \textbf{jigsaw operation} to perform style-content disentanglement and construct our style-texture pairs.

\textbf{Jigsaw Operation.}
As shown in~\Cref{fig:pipeline}, for a given reference image $I \in \mathbb{R}^{C \times H \times W}$, we first partition it into a grid of non-overlapping patches ${P_{i,j}}$, where each patch has size $S \times S$. These patches are then shuffled using a permutation function $\sigma$, which randomly reassembles them to disrupt structural semantics:
\begin{align}
    \label{eq:jigsaw-1}
    I_{\text{shuffled}} = \bigcup_{i,j} P_{\sigma(i,j)}, \quad \text{where each } P_{i,j} \in \mathbb{R}^{C \times S \times S}
\end{align}
This shuffling operation suppresses semantic information while preserving style attributes. We further apply stochastic masking with a mask ratio $p$ to control the proportion of patches that are masked:
\begin{align}
    \label{eq:jigsaw-2}
    I_{\text{jigsaw}} = \text{Mask} \left( I_{\text{shuffled}} \right) = \bigcup_{i,j} \left[ M_{i,j} \odot P_{\sigma(i,j)} + (1 - M_{i,j}) \odot \mu \right]
\end{align}
where $M_{i,j}$ is a binary mask with elements drawn from $\text{Bernoulli}(1-p)$, and $\mu$ is the masking background value. Similar to~\cite{he2022masked}, we encourage the remaining visible regions to reconstruct the styles of the masked patches.

During training, we use the jigsaw operation to process the current 3D object dataset to create style-texture pairs. For each 3D object, we render $K$ orthogonal views as texture targets, along with several additional random views as reference images. Each reference image is processed through the jigsaw operation to obtain $I_{\text{jigsaw}}$ as model input, while the original texture targets serve as ground truth supervision.
During inference, the jigsaw operation is applied to the user-provided reference image to produce the input for stylization, where only the shuffling operation is employed.

\subsection{Multi-view Style Generation}
\label{sec:pipeline}

After processing the reference image with the jigsaw operation, we create a large style-texture pair dataset and train a multi-view style generation model using this dataset. As illustrated in~\Cref{fig:pipeline}, the multi-view generation aims to produce multi-view consistent images, combining the geometric structure of \(M\) and the stylistic attributes of \(I\). 

\textbf{Geometric information injection}. To ensure the generated multiview images retain structural information with \(M\), we leverage geometric cues from  \(M\) and inject them into the denoising U-Net. Specifically, we first render both position and normal maps from the \(M\) from \(K\) predefined camera viewpoints, following the setup of~\cite{li2023sweetdreamer} and~\cite{bensadoun2024meta}. These maps are concatenated along the channel dimension to form the geometry condition \(\mathbf{G} \in \mathbb{R}^{B \times 2K \times H \times W}\), where the \(2K\) channels comprise \(K\) position maps and \(K\) normal maps.
Then the condition \(\mathbf{G} \) is processed by a trainable condition encoder \( \mathcal{G} \) based on T2I-Adapter~\citep{mou2024t2i}, which consists of a series of convolutional and downsampling layers. The resulting multi-scale features are injected directly into the corresponding scales of the Style U-Net denoiser through additive feature modulation, providing persistent geometric guidance throughout the denoising process.


\textbf{Style information injection}. To transfer style from the reference image \(I\), we first apply the jigsaw operation to suppress semantic information and disentangle style features, resulting in \(I_{\text{jigsaw}}\). This processed \(I_{\text{jigsaw}}\) is then encoded through a pre-trained VAE encoder \(\mathcal{E}\) to obtain latent features, which are fed into a pre-trained diffusion U-Net at timestep \(t = 0\). We extract intermediate hidden state features \(f_{\text{ref}}\) from the self-attention layers of this U-Net, which subsequently guide the stylization process in the style U-Net.
Our diffusion model employs a style U-Net architecture that incorporates a multi-branch attention block~\citep{huang2024mv} after each residual layer. These blocks consist of three parallel attention mechanisms: self-attention captures contextual relationships within each view; multi-view attention enforces consistency across different viewpoints using row-wise self-attention~\citep{li2024era3d}; and reference attention aligns the generation with the reference style by attending to \(f_{\text{ref}}\), as formalized below.

\textbf{Reference Attention.} We employ cross-attention for style transfer. In this module, the original input feature map \(f_{\text{in}}\) serves as the query, while \(f_{\text{ref}}\) serves as both the key and value. The reference attention operation is defined as:
\begin{align}
    \label{eq:ref-attn}
    \text{RefAttention}(f_{\text{in}}, f_{\text{ref}}) = \operatorname{softmax}\left( \frac{f_{\text{in}} f_{\text{ref}}^T}{\sqrt{d_k}} \right) f_{\text{ref}}
\end{align}
The softmax output represents relevance scores between the input features \(f_{\text{in}}\) and style features \(f_{\text{ref}}\), enabling dynamic style-aware recombination.
After computing the three attention outputs, the results are summed with the original input feature \(f_{\text{in}}\). The combined representation is further refined through a text-conditioned cross-attention layer. 
During training, ground-truth text captions with random dropout are used to improve robustness; during inference, generic prompts such as ``high quality'' are employed to maintain generalization and output quality.

\subsubsection{3D Style Baking}
3D Style Baking projects the pre-generated multi-view stylized images onto the UV texture space to produce fully textured 3D objects. This baking process consists of three main steps:~\textbf{Visibility-aware reprojection} establishes accurate pixel-to-UV correspondences while filtering occluded or invalid regions using camera and depth information; \textbf{3D inpainting} fills missing or invisible regions by computing a weighted average of the nearest neighboring pixels on the object surface. \textbf{Seamless composition} performs 2D inpainting in UV space to eliminate seam artifacts and ensure texture continuity.

\section{Experiments}

\textbf{Implementation Details.} Our approach is built upon Stable Diffusion XL~\citep{rombach2022high}. During training, we render each object from the Objaverse~\citep{deitke2023objaverse} to generate 6 orthogonal views as ground-truth and 4 random views as reference images. All images are scaled to a resolution of $512 \times 512$. In the jigsaw operation, the reference image is split into patches of size $64 \times 64$ during training and $128 \times 128$ during inference. A mask ratio between 0 and 0.25 achieves a balance between prediction capability and geometric consistency.
For model configuration, we apply a combined conditioning dropout strategy with a probability of 0.1, which independently drops the text condition, the image condition, or both simultaneously. The model is optimized using AdamW with a learning rate of $5 \times 10^{-5}$ for 10 epochs. We employ a DDPM sampler with 50 denoising steps during inference, with classifier-free guidance scale set to 3.0. Additionally, we adjust the log-SNR offset by $\log(n)$ where $n=6$ is the number of views.

\textbf{Evaluation Dataset.} For 3D objects, we select 20 objects from Objaverse~\citep{deitke2023objaverse} covering diverse categories, including both flat-surfaced and geometrically sharp shapes. Importantly, all selected meshes are distinct from those used during training to ensure a fair evaluation. For reference images, we first select style images from WikiArt~\citep{wikiart}, and additionally collect supplementary images manually from public sources. The \textbf{WikiArt dataset} includes 30 style images, with 5 examples each from 6 artistic genres: cityscape, figurative, flower painting, landscape, marina, and still-life. Furthermore, \textbf{our collected dataset} contains 40 extra images from the internet and existing publications to cover a broader spectrum of styles, such as Chinese ink painting, bronze/gold effects, Van Gogh-style art and cartoon illustrations. All images used comply with the Creative Commons Attribution 4.0 International (CC BY 4.0) license.

\textbf{Evaluation Metrics.} We employ several metrics to quantitatively evaluate performance between 6 orthogonal rendered views and the reference image. \textbf{Gram Matrix Similarity} and \textbf{AdaIN Distance} serve as style-fidelity measurements. Specifically, we extract style features from a pre-trained VGG-19 network. The Gram matrix similarity is calculated using the Frobenius norm between the style correlation matrices, and the AdaIN distance is computed as the sum of the $L_2$ norms between the feature means and standard deviations.
\textbf{CLIP Score} is used to measure style-content disentanglement. A lower CLIP score indicates weaker semantic correlation and more effective disentanglement of style information from the reference. The final metrics are reported as averages across multi-view computations.

\textbf{Baseline Methods.}
We compare our approach against several state-of-the-art methods in 3D texture/style generation. The selected baselines include two feed-forward generation methods~\citep{oztas20253d,huang2024mv} as well as one SDS-based optimization method~\citep{xie2024styletex}. For a fair comparison, all methods use the same reference image and object mesh.
Among baselines, \textbf{3D-style-LRM}~\citep{oztas20253d} generates initial multi-view images using InstantMesh~\citep{xu2024instantmesh} and fuses style information by blending attention outputs from both the original multi-view images and the style reference within the cross-attention module. We provide the required source images that maintain strict alignment with the input mesh.
\textbf{StyleTex}~\citep{xie2024styletex} disentangles style from reference images through orthogonal projection in the CLIP embedding space and guides texture generation iteratively using a diffusion-based style loss.
\textbf{MV-Adapter}~\citep{huang2024mv} employs a text-to-image diffusion model enhanced with adapter-based feature injection to produce multi-view consistent images with unified texture and style.

\subsection{Qualitative and Quantitative Comparisons}

\begin{figure}[t]
	\centering
        \includegraphics[width=0.98\textwidth]{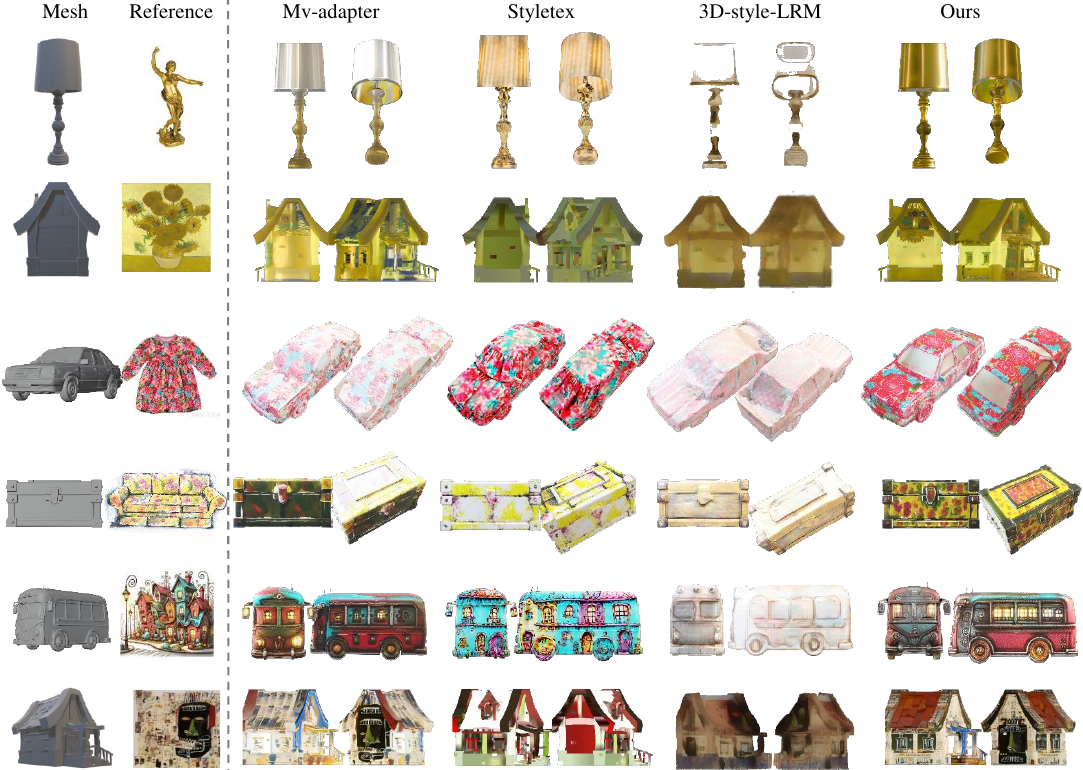} 
        \caption{\textbf{Qualitative comparison between 3D stylization methods on our collected dataset and WikiArt}. The left side of the dashed line displays the input object mesh and reference image. On the right, four groups of comparative results are shown, and each group has two selected viewpoints.}        
        \label{fig:experiment-full}
\end{figure}

\begin{table*}[t]
    \centering
    \resizebox{\linewidth}{!}{%
    \begin{tabular}{>{\centering\arraybackslash}m{5cm} ccc ccc >{\centering\arraybackslash}m{2.5cm}}
        \toprule
        \multirow{2}{*}{\textbf{Method}} & \multicolumn{3}{c}{\textbf{Collected Data}} & \multicolumn{3}{c}{\textbf{WikiArt}} & \multirow{2}{*}{\textbf{Cost Time}} \\
        & \textbf{Gram} $\downarrow$ & \textbf{AdaIN} $\downarrow$ & \textbf{CLIP} $\downarrow$ & \textbf{Gram} $\downarrow$ & \textbf{AdaIN} $\downarrow$ & \textbf{CLIP} $\downarrow$ & \\
        \midrule
        StyleTex (TOG 2025) & 5.35 & 124.54 & \textcolor{red}{\textbf{0.205}} & 6.54 & 149.27 & \textcolor{red}{\textbf{0.208}} & 15min \\
        MV-Adapter (ICCV 2025) & \textcolor{blue}{4.85} & \textcolor{blue}{114.04} & 0.214 & \textcolor{blue}{4.91} & \textcolor{blue}{122.19} & 0.213 & $\sim$40\text{s} \\
        3D-style-LRM (SIGGRAPH 2025) & 5.78 & 136.22 & 0.215 & 5.49 & 139.86 & \textcolor{blue}{0.210} & $\sim$35\text{s} \\
        \cmidrule(lr){1-8}
        \textbf{ours} & \textcolor{red}{\textbf{4.81}} & \textcolor{red}{\textbf{113.38}} & \textcolor{blue}{0.213} & \textcolor{red}{\textbf{4.82}} & \textcolor{red}{\textbf{120.54}} &  \textcolor{blue}{0.210} & $\sim$40\text{s} \\
        \bottomrule
    \end{tabular}%
    }
    \caption{\textbf{Quantitative comparison between 3D stylization methods on our collected dataset and WikiArt.} Lower Gram and AdaIN value reflect better style fidelity, and lower CLIP score reflects more effective style-content disentanglement.}
    \label{tab:main_results}
\end{table*}

\textbf{Qualitative Comparison.}~\Cref{fig:experiment-full} presents a qualitative comparison of different methods for transferring reference styles onto geometric meshes.
The results show that 3D-Style-LRM exhibits significant limitations in preserving the geometric fidelity of the original mesh, resulting in inconsistent surfaces.
All baseline methods suffer from style infidelity compared to the reference image, such as color shifts in the clothing (row 3) or loss of texture patterns in the sofa (row 4). Additionally, MV-Adapter incorrectly transfers the entire texture map layout onto the target object.
In contrast, our method demonstrates superior visual quality, with color distribution and texture details consistent with the reference. Furthermore, our disentanglement and recombination strategy effectively applies style attributes to semantically appropriate structures. As shown in row 2, our approach applies a floral pattern to the roof of a building while maintaining pure-colored walls, enabling the assignment of distinct styles to different structural components.

\textbf{Quantitative Comparison.}~\Cref{tab:main_results} presents a quantitative comparison of 3D stylization methods. As demonstrated, our method achieves significantly superior performance on style-related metrics including Gram matrix similarity and AdaIN, indicating exceptional style consistency with the reference and multi-view coherence. Furthermore, our approach attains competitive CLIP scores, second only to StyleTex, which effectively demonstrates successful semantic disentanglement. The slightly lower CLIP performance compared to StyleTex can be attributed to the fact that StyleTex utilizes style text descriptions as additional input conditions, providing ground-truth information. Besides, compared to SDS-based methods like StyleTex, our approach is more efficient in terms of computational time.

\subsection{Ablation Study}
\begin{figure}[t]
\centering
\includegraphics[width=0.98\textwidth]{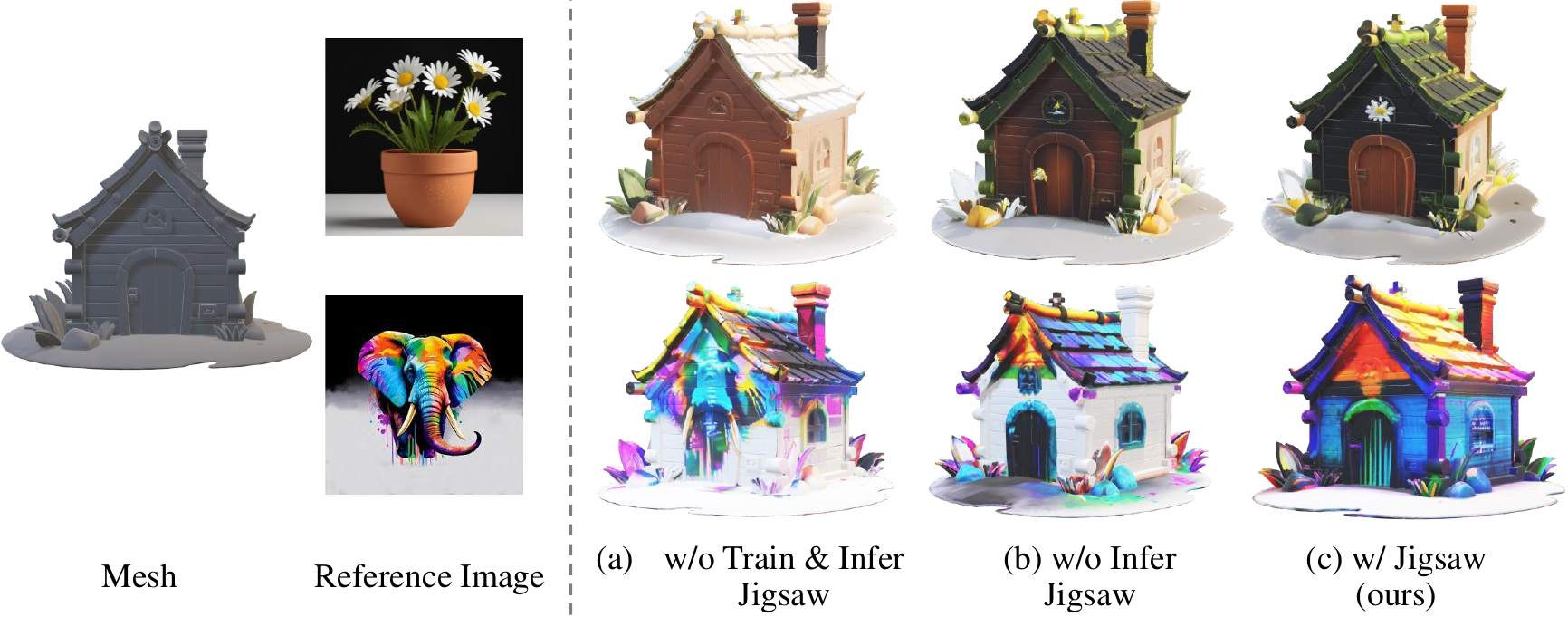}
\caption{
\textbf{Ablation study on the Jigsaw module.}
The left side shows the input object mesh and reference style image.
The right side presents groups of stylization results under different Jigsaw settings:
\textbf{(a) w/o Train \& Infer Jigsaw}: training and reference process without jigsaw operation;
\textbf{(b) w/o Infer Jigsaw}: only inference process without jigsaw operation;
\textbf{(c) w/ Train \& Infer Jigsaw (Ours)}: our approach applies the jigsaw operation in both training and inference phases.
}
\label{fig:ablation-jigsaw}
\end{figure}

\textbf{Ablation Study on Jigsaw Module.}~\Cref{fig:ablation-jigsaw} presents an ablation study evaluating the effectiveness of the jigsaw module under different configurations.
Figure (a) shows that without the jigsaw operation, the first row lacks the flower style, while the second row exhibits semantic entanglement of the elephant.
When applying the jigsaw module during training as shown in Figure (b), the flower style is attributed to the house roof and the elephant semantics are compressed.
Furthermore, applying jigsaw during both training and inference, as shown in Figure (c), preserves better style fidelity to the reference images, as demonstrated by the color distribution of the elephant and the detailed floral patterns on the house.

We further provide additional ablation studies, including those on the \textbf{patch size $S$} and \textbf{mask ratio $p$} in the jigsaw operation, in the~\Cref{supp-ablation-study}.

\subsection{More Applications and Limitations}
\begin{figure}[t]
\centering
\includegraphics[width=0.98\textwidth]{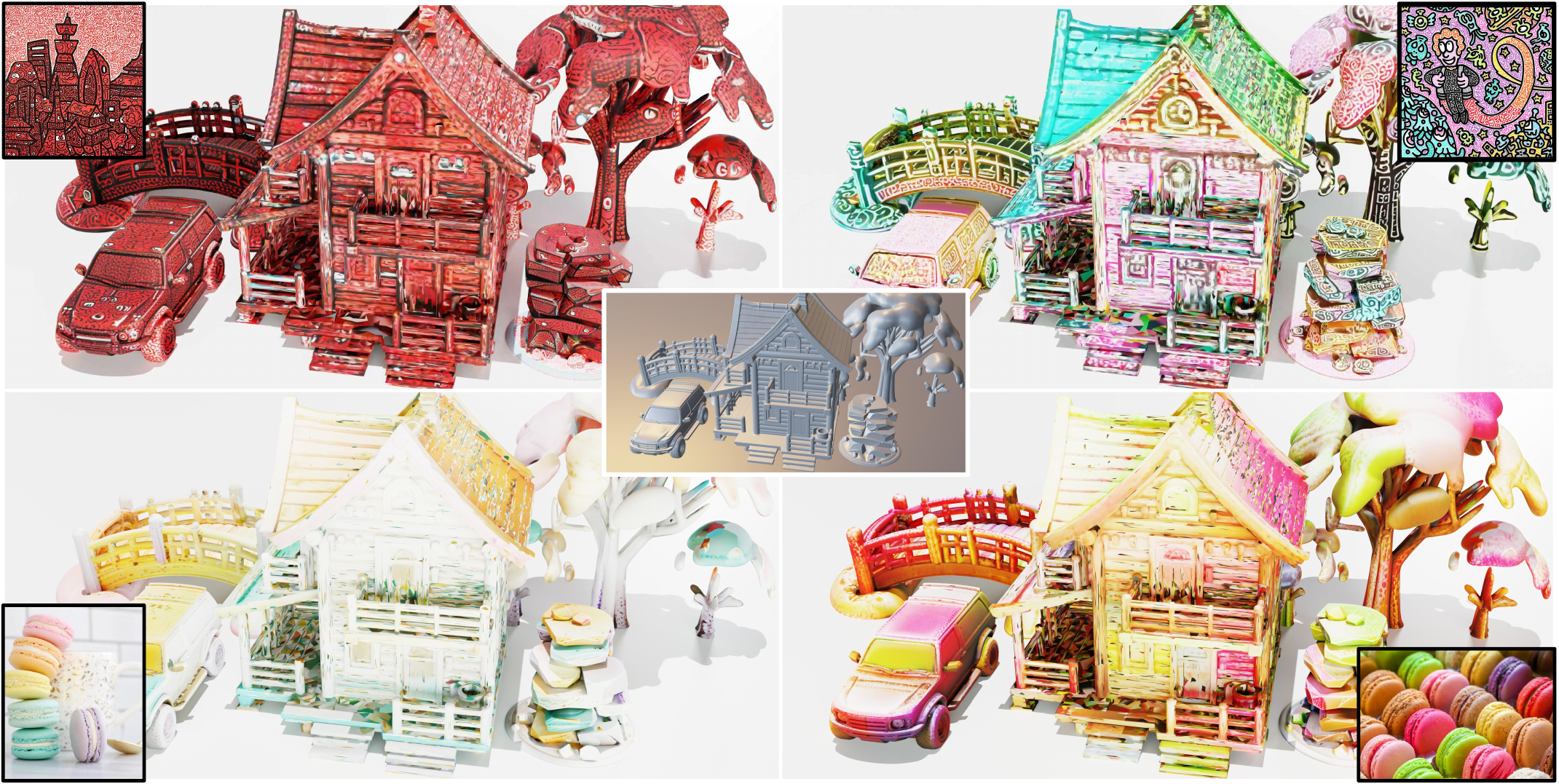}
\caption{
The first figure demonstrates strong geometric awareness, as the generated \textbf{sketch lines align precisely with the 3D feature lines of the objects}. Across all four figures, local regions also exhibit reasonable style attribution, as the house roof and wall maintain color consistency with the reference images.
}
\label{fig:more-application-1}
\end{figure}

\textbf{Multiple Objects Scene Stylization}. We collected several scene-related 3D objects and applied our method to stylize them. As shown in \Cref{fig:more-application-1}, our approach successfully achieves consistent style coherence both within individual objects and across different objects in the scene.

We further demonstrate other applications, including \textbf{tileable texture generation} and \textbf{partial stylization}, in the appendix~\Cref{supp-more-application}.

\begin{figure}[t]
\centering
\includegraphics[width=0.98\textwidth]{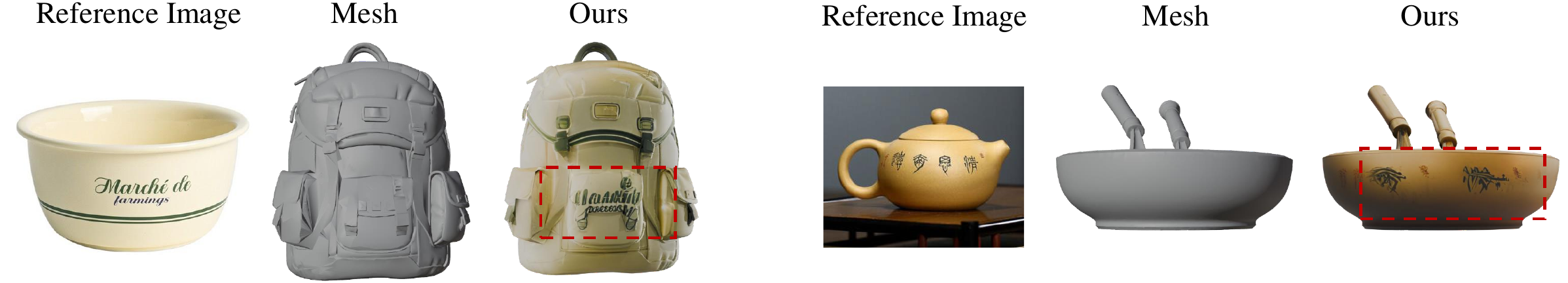}
\caption{
\textbf{Limitations of our method.} The dashed red boxes indicate failure cases in style transfer, such as text or symbolic patterns that do not match the reference image.
}
\label{fig:limitation}
\end{figure}
 
\textbf{Limitations.} During the style transfer process, our method struggles to accurately preserve fine-grained patterns such as text or symbols, as shown in~\Cref{fig:limitation}. This limitation is attributable to Stable Diffusion backbone (SDXL) we used, which lacks the capability to reliably generate or reconstruct precise textual and symbolic structures.

\section{Conclusion}
In this paper, we propose a new framework to transfer the style of a reference image to a 3D object while preserving its geometric structure. 
We propose a novel framework centered around the jigsaw operation to create a style-3D pair dataset. 
We apply a multiview generation pipeline to utilize the extracted style features from the jigsawed images.
Geometric cues such as normal and position maps are further incorporated to enhance structural alignment.
Extensive experiments demonstrate that our method achieves state-of-the-art performance across several 3D stylization benchmarks and shows strong generalization to downstream tasks such as partial object stylization, multi-object scene styling, and tileable texture generation.

\newpage

\textbf{Ethical Statement.} Our work presents a method for 3D style transfer using multi-view diffusion models and a jigsaw-based disentanglement mechanism. We acknowledge that our approach could potentially be misused to create misleading or harmful content, such as generating stylized 3D objects that infringe upon intellectual property, appropriate cultural or artistic styles without permission, or propagate visual misinformation.
We strongly emphasize that the use of this technology should adhere to the relevant legal frameworks and community standards. All 3D assets and reference images should be properly licensed or used in accordance with fair use principles.

\textbf{Reproducibility Statement.} To facilitate the reproducibility of our work, we provide code in the supplementary materials. A detailed README document is included, which covers the environment configuration and instructions for loading pre-trained weights.

\newpage

\bibliography{iclr2026_conference}

\begin{thebibliography}{42}
\providecommand{\natexlab}[1]{#1}
\providecommand{\url}[1]{\texttt{#1}}
\expandafter\ifx\csname urlstyle\endcsname\relax
  \providecommand{\doi}[1]{doi: #1}\else
  \providecommand{\doi}{doi: \begingroup \urlstyle{rm}\Url}\fi

\bibitem[Bensadoun et~al.(2024)Bensadoun, Kleiman, Azuri, Harosh, Vedaldi, Neverova, and Gafni]{bensadoun2024meta}
Raphael Bensadoun, Yanir Kleiman, Idan Azuri, Omri Harosh, Andrea Vedaldi, Natalia Neverova, and Oran Gafni.
\newblock Meta 3d texturegen: Fast and consistent texture generation for 3d objects.
\newblock \emph{arXiv preprint arXiv:2407.02430}, 2024.

\bibitem[Chang et~al.(2015)Chang, Funkhouser, Guibas, Hanrahan, Huang, Li, Savarese, Savva, Song, Su, et~al.]{chang2015shapenet}
Angel~X Chang, Thomas Funkhouser, Leonidas Guibas, Pat Hanrahan, Qixing Huang, Zimo Li, Silvio Savarese, Manolis Savva, Shuran Song, Hao Su, et~al.
\newblock Shapenet: An information-rich 3d model repository.
\newblock \emph{arXiv preprint arXiv:1512.03012}, 2015.

\bibitem[Deitke et~al.(2023{\natexlab{a}})Deitke, Liu, Wallingford, Ngo, Michel, Kusupati, Fan, Laforte, Voleti, Gadre, et~al.]{deitke2023objaverse-xl}
Matt Deitke, Ruoshi Liu, Matthew Wallingford, Huong Ngo, Oscar Michel, Aditya Kusupati, Alan Fan, Christian Laforte, Vikram Voleti, Samir~Yitzhak Gadre, et~al.
\newblock Objaverse-xl: A universe of 10m+ 3d objects.
\newblock \emph{Advances in Neural Information Processing Systems}, 36:\penalty0 35799--35813, 2023{\natexlab{a}}.

\bibitem[Deitke et~al.(2023{\natexlab{b}})Deitke, Schwenk, Salvador, Weihs, Michel, VanderBilt, Schmidt, Ehsani, Kembhavi, and Farhadi]{deitke2023objaverse}
Matt Deitke, Dustin Schwenk, Jordi Salvador, Luca Weihs, Oscar Michel, Eli VanderBilt, Ludwig Schmidt, Kiana Ehsani, Aniruddha Kembhavi, and Ali Farhadi.
\newblock Objaverse: A universe of annotated 3d objects.
\newblock In \emph{Proceedings of the IEEE/CVF conference on computer vision and pattern recognition}, pp.\  13142--13153, 2023{\natexlab{b}}.

\bibitem[Frenkel et~al.(2024)Frenkel, Vinker, Shamir, and Cohen-Or]{frenkel2024implicit}
Yarden Frenkel, Yael Vinker, Ariel Shamir, and Daniel Cohen-Or.
\newblock Implicit style-content separation using b-lora.
\newblock In \emph{European Conference on Computer Vision}, pp.\  181--198. Springer, 2024.

\bibitem[Fujiwara et~al.(2024)Fujiwara, Mukuta, and Harada]{fujiwara2024style}
Haruo Fujiwara, Yusuke Mukuta, and Tatsuya Harada.
\newblock Style-nerf2nerf: 3d style transfer from style-aligned multi-view images.
\newblock In \emph{SIGGRAPH Asia 2024 Conference Papers}, pp.\  1--10, 2024.

\bibitem[Gatys et~al.(2016)Gatys, Ecker, and Bethge]{gatys2016image}
Leon~A Gatys, Alexander~S Ecker, and Matthias Bethge.
\newblock Image style transfer using convolutional neural networks.
\newblock In \emph{Proceedings of the IEEE conference on computer vision and pattern recognition}, pp.\  2414--2423, 2016.

\bibitem[Gu et~al.(2018)Gu, Chen, Liao, and Yuan]{gu2018arbitrary}
Shuyang Gu, Congliang Chen, Jing Liao, and Lu~Yuan.
\newblock Arbitrary style transfer with deep feature reshuffle.
\newblock In \emph{Proceedings of the IEEE conference on computer vision and pattern recognition}, pp.\  8222--8231, 2018.

\bibitem[He et~al.(2022)He, Chen, Xie, Li, Doll{\'a}r, and Girshick]{he2022masked}
Kaiming He, Xinlei Chen, Saining Xie, Yanghao Li, Piotr Doll{\'a}r, and Ross Girshick.
\newblock Masked autoencoders are scalable vision learners.
\newblock In \emph{Proceedings of the IEEE/CVF conference on computer vision and pattern recognition}, pp.\  16000--16009, 2022.

\bibitem[Hertz et~al.(2024)Hertz, Voynov, Fruchter, and Cohen-Or]{hertz2024style}
Amir Hertz, Andrey Voynov, Shlomi Fruchter, and Daniel Cohen-Or.
\newblock Style aligned image generation via shared attention.
\newblock In \emph{Proceedings of the IEEE/CVF Conference on Computer Vision and Pattern Recognition}, pp.\  4775--4785, 2024.

\bibitem[Hu et~al.(2022)Hu, Shen, Wallis, Allen-Zhu, Li, Wang, Wang, Chen, et~al.]{hu2022lora}
Edward~J Hu, Yelong Shen, Phillip Wallis, Zeyuan Allen-Zhu, Yuanzhi Li, Shean Wang, Lu~Wang, Weizhu Chen, et~al.
\newblock Lora: Low-rank adaptation of large language models.
\newblock \emph{ICLR}, 1\penalty0 (2):\penalty0 3, 2022.

\bibitem[Huang \& Belongie(2017)Huang and Belongie]{huang2017arbitrary}
Xun Huang and Serge Belongie.
\newblock Arbitrary style transfer in real-time with adaptive instance normalization.
\newblock In \emph{Proceedings of the IEEE international conference on computer vision}, pp.\  1501--1510, 2017.

\bibitem[Huang et~al.(2024)Huang, Guo, Wang, Yi, Ma, Cao, and Sheng]{huang2024mv}
Zehuan Huang, Yuan-Chen Guo, Haoran Wang, Ran Yi, Lizhuang Ma, Yan-Pei Cao, and Lu~Sheng.
\newblock Mv-adapter: Multi-view consistent image generation made easy.
\newblock \emph{arXiv preprint arXiv:2412.03632}, 2024.

\bibitem[Jeong et~al.(2024)Jeong, Kim, Choi, Lee, and Uh]{jeong2024visual}
Jaeseok Jeong, Junho Kim, Yunjey Choi, Gayoung Lee, and Youngjung Uh.
\newblock Visual style prompting with swapping self-attention.
\newblock \emph{arXiv preprint arXiv:2402.12974}, 2024.

\bibitem[Li et~al.(2024)Li, Liu, Long, Zhang, Lin, Li, Qi, Zhang, Xue, Luo, et~al.]{li2024era3d}
Peng Li, Yuan Liu, Xiaoxiao Long, Feihu Zhang, Cheng Lin, Mengfei Li, Xingqun Qi, Shanghang Zhang, Wei Xue, Wenhan Luo, et~al.
\newblock Era3d: High-resolution multiview diffusion using efficient row-wise attention.
\newblock \emph{Advances in Neural Information Processing Systems}, 37:\penalty0 55975--56000, 2024.

\bibitem[Li et~al.(2023)Li, Chen, Chen, and Tan]{li2023sweetdreamer}
Weiyu Li, Rui Chen, Xuelin Chen, and Ping Tan.
\newblock Sweetdreamer: Aligning geometric priors in 2d diffusion for consistent text-to-3d.
\newblock \emph{arXiv preprint arXiv:2310.02596}, 2023.

\bibitem[Li et~al.(2025)Li, Wang, Zheng, Luo, and Wen]{li2025sparc3d}
Zhihao Li, Yufei Wang, Heliang Zheng, Yihao Luo, and Bihan Wen.
\newblock Sparc3d: Sparse representation and construction for high-resolution 3d shapes modeling.
\newblock \emph{arXiv preprint arXiv:2505.14521}, 2025.

\bibitem[Lin et~al.(2023)Lin, Gao, Tang, Takikawa, Zeng, Huang, Kreis, Fidler, Liu, and Lin]{lin2023magic3d}
Chia-Hao Lin, Jun Gao, Luming Tang, Towaki Takikawa, Xiaohui Zeng, Xun Huang, Karsten Kreis, Sanja Fidler, Ming-Yu Liu, and Tsung-Yi Lin.
\newblock Magic3d: High-resolution text-to-3d content creation.
\newblock In \emph{Proceedings of the IEEE/CVF Conference on Computer Vision and Pattern Recognition}, pp.\  300--309, 2023.

\bibitem[Liu et~al.(2023{\natexlab{a}})Liu, Wu, Van~Hoorick, Tokmakov, Vondrick, and Fathi]{liu2023zero}
Ruoshi Liu, Rundi Wu, Basile Van~Hoorick, Pavel Tokmakov, Carl Vondrick, and Ali Fathi.
\newblock Zero-1-to-3: Zero-shot one image to 3d object.
\newblock In \emph{Proceedings of the IEEE/CVF International Conference on Computer Vision}, pp.\  22026--22037, 2023{\natexlab{a}}.

\bibitem[Liu et~al.(2023{\natexlab{b}})Liu, Lin, Zeng, Long, Liu, Komura, and Wang]{liu2023syncdreamer}
Yuan Liu, Cheng Lin, Zijiao Zeng, Xiaoxiao Long, Lingjie Liu, Taku Komura, and Wenping Wang.
\newblock Syncdreamer: Generating multiview-consistent images from a single-view image.
\newblock \emph{arXiv preprint arXiv:2309.03453}, 2023{\natexlab{b}}.

\bibitem[Lu et~al.(2019)Lu, Zhao, Yao, Chen, Xu, and Zhang]{lu2019closed}
Ming Lu, Hao Zhao, Anbang Yao, Yurong Chen, Feng Xu, and Li~Zhang.
\newblock A closed-form solution to universal style transfer.
\newblock In \emph{Proceedings of the IEEE/CVF international conference on computer vision}, pp.\  5952--5961, 2019.

\bibitem[Mescheder et~al.(2019)Mescheder, Oechsle, Niemeyer, Nowozin, and Geiger]{mescheder2019occupancy}
Lars Mescheder, Michael Oechsle, Michael Niemeyer, Sebastian Nowozin, and Andreas Geiger.
\newblock Occupancy networks: Learning 3d reconstruction in function space.
\newblock In \emph{Proceedings of the IEEE/CVF conference on computer vision and pattern recognition}, pp.\  4460--4470, 2019.

\bibitem[Mildenhall et~al.(2021)Mildenhall, Srinivasan, Tancik, Barron, Ramamoorthi, and Ng]{mildenhall2021nerf}
Ben Mildenhall, Pratul~P Srinivasan, Matthew Tancik, Jonathan~T Barron, Ravi Ramamoorthi, and Ren Ng.
\newblock Nerf: Representing scenes as neural radiance fields for view synthesis.
\newblock \emph{Communications of the ACM}, 65\penalty0 (1):\penalty0 99--106, 2021.

\bibitem[Mou et~al.(2024)Mou, Wang, Xie, Wu, Zhang, Qi, and Shan]{mou2024t2i}
Chong Mou, Xintao Wang, Liangbin Xie, Yanze Wu, Jian Zhang, Zhongang Qi, and Ying Shan.
\newblock T2i-adapter: Learning adapters to dig out more controllable ability for text-to-image diffusion models.
\newblock In \emph{Proceedings of the AAAI conference on artificial intelligence}, volume~38, pp.\  4296--4304, 2024.

\bibitem[Oztas et~al.(2025)Oztas, Ceylan, and Dundar]{oztas20253d}
Ipek Oztas, Duygu Ceylan, and Aysegul Dundar.
\newblock 3d stylization via large reconstruction model.
\newblock In \emph{Proceedings of the Special Interest Group on Computer Graphics and Interactive Techniques Conference Conference Papers}, pp.\  1--11, 2025.

\bibitem[Park et~al.(2019)Park, Florence, Straub, Newcombe, and Lovegrove]{park2019deepsdf}
Jeong~Joon Park, Peter Florence, Julian Straub, Richard Newcombe, and Steven Lovegrove.
\newblock Deepsdf: Learning continuous signed distance functions for shape representation.
\newblock In \emph{Proceedings of the IEEE/CVF conference on computer vision and pattern recognition}, pp.\  165--174, 2019.

\bibitem[Poole et~al.(2022)Poole, Jain, Barron, and Mildenhall]{poole2022dreamfusion}
Ben Poole, Ajay Jain, Jonathan~T Barron, and Ben Mildenhall.
\newblock Dreamfusion: Text-to-3d using 2d diffusion.
\newblock \emph{arXiv preprint arXiv:2209.14988}, 2022.

\bibitem[Rombach et~al.(2022)Rombach, Blattmann, Lorenz, Esser, and Ommer]{rombach2022high}
Robin Rombach, Andreas Blattmann, Dominik Lorenz, Patrick Esser, and Bj{\"o}rn Ommer.
\newblock High-resolution image synthesis with latent diffusion models.
\newblock In \emph{Proceedings of the IEEE/CVF Conference on Computer Vision and Pattern Recognition}, pp.\  10684--10695, 2022.

\bibitem[Shi et~al.(2023)Shi, Wang, Ye, Mai, Li, and Yang]{shi2023MVDream}
Yichun Shi, Peng Wang, Jianglong Ye, Long Mai, Kejie Li, and Xiao Yang.
\newblock Mvdream: Multi-view diffusion for 3d generation.
\newblock \emph{arXiv preprint arXiv:2308.16512}, 2023.

\bibitem[Song et~al.(2024)Song, Huang, Xie, Wang, and Wang]{song2024style3d}
Bingjie Song, Xin Huang, Ruting Xie, Xue Wang, and Qing Wang.
\newblock Style3d: Attention-guided multi-view style transfer for 3d object generation.
\newblock \emph{arXiv preprint arXiv:2412.03571}, 2024.

\bibitem[Wang et~al.(2023)Wang, Wang, Xie, Qi, Shan, Wang, and Luo]{wang2023styleadapter}
Zhouxia Wang, Xintao Wang, Liangbin Xie, Zhongang Qi, Ying Shan, Wenping Wang, and Ping Luo.
\newblock Styleadapter: A unified stylized image generation model.
\newblock \emph{arXiv preprint arXiv:2309.01770}, 2023.

\bibitem[{WikiArt}(2014)]{wikiart}
{WikiArt}.
\newblock The online visual art encyclopedia.
\newblock \url{https://www.wikiart.org/}, 2014.

\bibitem[Wu et~al.(2024)Wu, Lin, Zhang, Zeng, Xu, Torr, Cao, and Yao]{wu2024direct3d}
Shuang Wu, Youtian Lin, Feihu Zhang, Yifei Zeng, Jingxi Xu, Philip Torr, Xun Cao, and Yao Yao.
\newblock Direct3d: Scalable image-to-3d generation via 3d latent diffusion transformer.
\newblock \emph{Advances in Neural Information Processing Systems}, 37:\penalty0 121859--121881, 2024.

\bibitem[Xiang et~al.(2025)Xiang, Lv, Xu, Deng, Wang, Zhang, Chen, Tong, and Yang]{xiang2025structured}
Jianfeng Xiang, Zelong Lv, Sicheng Xu, Yu~Deng, Ruicheng Wang, Bowen Zhang, Dong Chen, Xin Tong, and Jiaolong Yang.
\newblock Structured 3d latents for scalable and versatile 3d generation.
\newblock In \emph{Proceedings of the Computer Vision and Pattern Recognition Conference}, pp.\  21469--21480, 2025.

\bibitem[Xie et~al.(2024)Xie, Zhang, Tang, Wu, Chen, Li, and Jin]{xie2024styletex}
Zhiyu Xie, Yuqing Zhang, Xiangjun Tang, Yiqian Wu, Dehan Chen, Gongsheng Li, and Xiaogang Jin.
\newblock Styletex: Style image-guided texture generation for 3d models.
\newblock \emph{ACM Transactions on Graphics (TOG)}, 43\penalty0 (6):\penalty0 1--14, 2024.

\bibitem[Xu et~al.(2024)Xu, Cheng, Gao, Wang, Gao, and Shan]{xu2024instantmesh}
Jiale Xu, Weihao Cheng, Yiming Gao, Xintao Wang, Shenghua Gao, and Ying Shan.
\newblock Instantmesh: Efficient 3d mesh generation from a single image with sparse-view large reconstruction models.
\newblock \emph{arXiv preprint arXiv:2404.07191}, 2024.

\bibitem[Ye et~al.(2023)Ye, Zhang, Liu, Han, and Yang]{ye2023ip}
Hu~Ye, Jun Zhang, Sibo Liu, Xiao Han, and Wei Yang.
\newblock Ip-adapter: Text compatible image prompt adapter for text-to-image diffusion models.
\newblock \emph{arXiv preprint arXiv:2308.06721}, 2023.

\bibitem[Zhang et~al.(2023{\natexlab{a}})Zhang, Tang, Niessner, and Wonka]{zhang20233dshape2vecset}
Biao Zhang, Jiapeng Tang, Matthias Niessner, and Peter Wonka.
\newblock 3dshape2vecset: A 3d shape representation for neural fields and generative diffusion models.
\newblock \emph{ACM Transactions On Graphics (TOG)}, 42\penalty0 (4):\penalty0 1--16, 2023{\natexlab{a}}.

\bibitem[Zhang et~al.(2024)Zhang, Wang, Zhang, Qiu, Pang, Jiang, Yang, Xu, and Yu]{zhang2024clay}
Longwen Zhang, Ziyu Wang, Qixuan Zhang, Qiwei Qiu, Anqi Pang, Haoran Jiang, Wei Yang, Lan Xu, and Jingyi Yu.
\newblock Clay: A controllable large-scale generative model for creating high-quality 3d assets.
\newblock \emph{ACM Transactions on Graphics (TOG)}, 43\penalty0 (4):\penalty0 1--20, 2024.

\bibitem[Zhang et~al.(2023{\natexlab{b}})Zhang, Rao, and Agrawala]{zhang2023adding}
Lvmin Zhang, Anyi Rao, and Maneesh Agrawala.
\newblock Adding conditional control to text-to-image diffusion models.
\newblock In \emph{Proceedings of the IEEE/CVF international conference on computer vision}, pp.\  3836--3847, 2023{\natexlab{b}}.

\bibitem[Zhang et~al.(2022)Zhang, Tang, Dong, Huang, Ma, Lee, and Xu]{zhang2022domain}
Yuxin Zhang, Fan Tang, Weiming Dong, Haibin Huang, Chongyang Ma, Tong-Yee Lee, and Changsheng Xu.
\newblock Domain enhanced arbitrary image style transfer via contrastive learning.
\newblock In \emph{ACM SIGGRAPH 2022 conference proceedings}, pp.\  1--8, 2022.

\bibitem[Zhao et~al.(2025)Zhao, Lai, Lin, Zhao, Liu, Yang, Feng, Yang, Zhang, Yang, et~al.]{zhao2025hunyuan3d}
Zibo Zhao, Zeqiang Lai, Qingxiang Lin, Yunfei Zhao, Haolin Liu, Shuhui Yang, Yifei Feng, Mingxin Yang, Sheng Zhang, Xianghui Yang, et~al.
\newblock Hunyuan3d 2.0: Scaling diffusion models for high resolution textured 3d assets generation.
\newblock \emph{arXiv preprint arXiv:2501.12202}, 2025.

\end{thebibliography}
\bibliographystyle{iclr2026_conference}

\newpage

\appendix
\section{Appendix}

\subsection{Use of Large Language Models (LLMs)}
We declare that Large Language Models (LLMs) were used solely as an auxiliary tool in the writing process of this paper, specifically for tasks such as checking and correcting grammatical errors and ensuring consistency in formatting and terminology. We emphasize that the core ideas, theoretical derivations, experimental design, and result analysis were entirely conceived and conducted by the authors.

\subsection{Proof of Style Statistics Preservation under Shuffle Operation}
Let \( I \in \mathbb{R}^{3 \times H \times W} \) denote an input image with 3 channels, height \( H \), and width \( W \). We partition \( I \) into \( N \times N \) non-overlapping patches \( \{P_{i,j}\} \), each of size \( \frac{H}{N} \times \frac{W}{N} \). Let \( I_{\text{shuffled}} \) be the image after applying a permutation \( \sigma \) to these patches:

\[
I_{\text{shuffled}} = \bigcup_{i,j} P_{\sigma(i,j)}
\]

\textbf{Mean Preservation.} The mean of channel \( c \) of the original image is:
\[
\mu_c = \frac{1}{HW} \sum_{i=1}^{H} \sum_{j=1}^{W} I(c, i, j)
\]
After shuffling, only the spatial positions of pixel values are permuted. Therefore, the sum over all positions remains identical:
\[
\sum_{i=1}^{H} \sum_{j=1}^{W} I(c, i, j) = \sum_{i=1}^{H} \sum_{j=1}^{W} I_{\text{shuffled}}(c, i, j)
\]
Thus,
\[
\mu_c' = \frac{1}{HW} \sum_{i=1}^{H} \sum_{j=1}^{W} I_{\text{shuffled}}(c, i, j) = \frac{1}{HW} \sum_{i=1}^{H} \sum_{j=1}^{W} I(c, i, j) = \mu_c
\]

\textbf{Variance Preservation.} The variance of channel \( c \) is defined as:
\[
\sigma_c^2 = \frac{1}{HW} \sum_{i=1}^{H} \sum_{j=1}^{W} \left( I(c, i, j) - \mu_c \right)^2
\]
Since both the pixel values \( I(c,i,j) \) and the mean \( \mu_c \) remain unchanged under shuffling, each squared term \( (I(c,i,j) - \mu_c)^2 \) is preserved. Therefore, the sum of squared deviations remains the same:
\[
\sum_{i=1}^{H} \sum_{j=1}^{W} \left( I(c, i, j) - \mu_c \right)^2 = \sum_{i=1}^{H} \sum_{j=1}^{W} \left( I_{\text{shuffled}}(c, i, j) - \mu_c' \right)^2
\]
Hence,
\[
(\sigma_c^2)' = \sigma_c^2
\]

According to the above, shuffling patches changes the spatial arrangement of pixels but does not alter their first-order (mean) or second-order (variance) statistics.

\subsection{Style Consistency Metrics.}  
To quantitatively evaluate style consistency between generated multi-view images and the reference style image, we employ two widely adopted metrics in style transfer: \textbf{Gram Matrix Similarity} and \textbf{AdaIN Distance}. Features are extracted from five key ReLU layers (1\_1, 2\_1, 3\_1, 4\_1, 5\_1) of a pre-trained VGG-19 network.

\noindent
\textbf{Gram Matrix Similarity} captures the correlations between feature channels. For a feature map \( \mathbf{F} \in \mathbb{R}^{C \times H \times W} \), the Gram matrix \( \mathbf{G} \in \mathbb{R}^{C \times C} \) is computed as:
\[
\mathbf{G} = \frac{1}{C \cdot H \cdot W} \mathbf{F} \mathbf{F}^\top
\]
The style similarity between the reference image and a generated view is measured using the Frobenius norm:
\[
\mathcal{L}_{\text{Gram}} = \| \mathbf{G}_{\text{ref}} - \mathbf{G}_{\text{gen}} \|_F
\]

\noindent
\textbf{AdaIN Distance} measures the discrepancy in first-order (mean) and second-order (standard deviation) statistics of deep features:
\[
\mathcal{L}_{\mu} = \| \mu(\mathbf{F}_{\text{ref}}) - \mu(\mathbf{F}_{\text{gen}}) \|_2, \quad
\mathcal{L}_{\sigma} = \| \sigma(\mathbf{F}_{\text{ref}}) - \sigma(\mathbf{F}_{\text{gen}}) \|_2
\]
\[
\mathcal{L}_{\text{AdaIN}} = \mathcal{L}_{\mu} + \mathcal{L}_{\sigma}
\]

Both metrics are computed across five VGG-19 layers and averaged over all views. Lower values indicate better style consistency.

\subsection{More Results}  

\begin{figure}[t]
\centering
\includegraphics[width=0.98\textwidth]{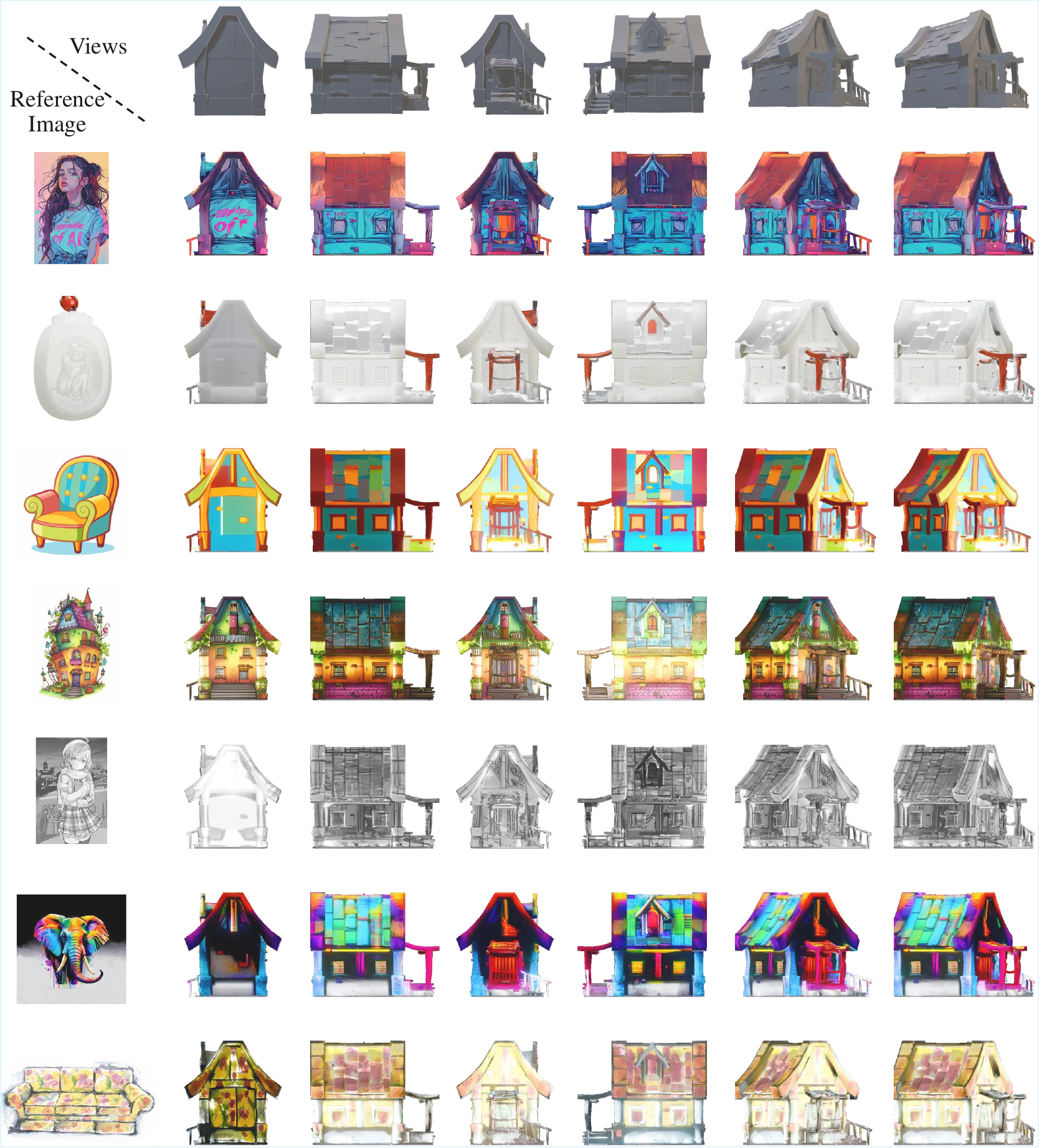}
\caption{
\textbf{More 3D Stylization Results of our method.}  Multi-view rendering results by applying diverse reference style images to a ``house'' object.
}
\label{supp:collected_data_more_result_part_1}
\end{figure}

\begin{figure}[t]
\centering
\includegraphics[width=0.98\textwidth]{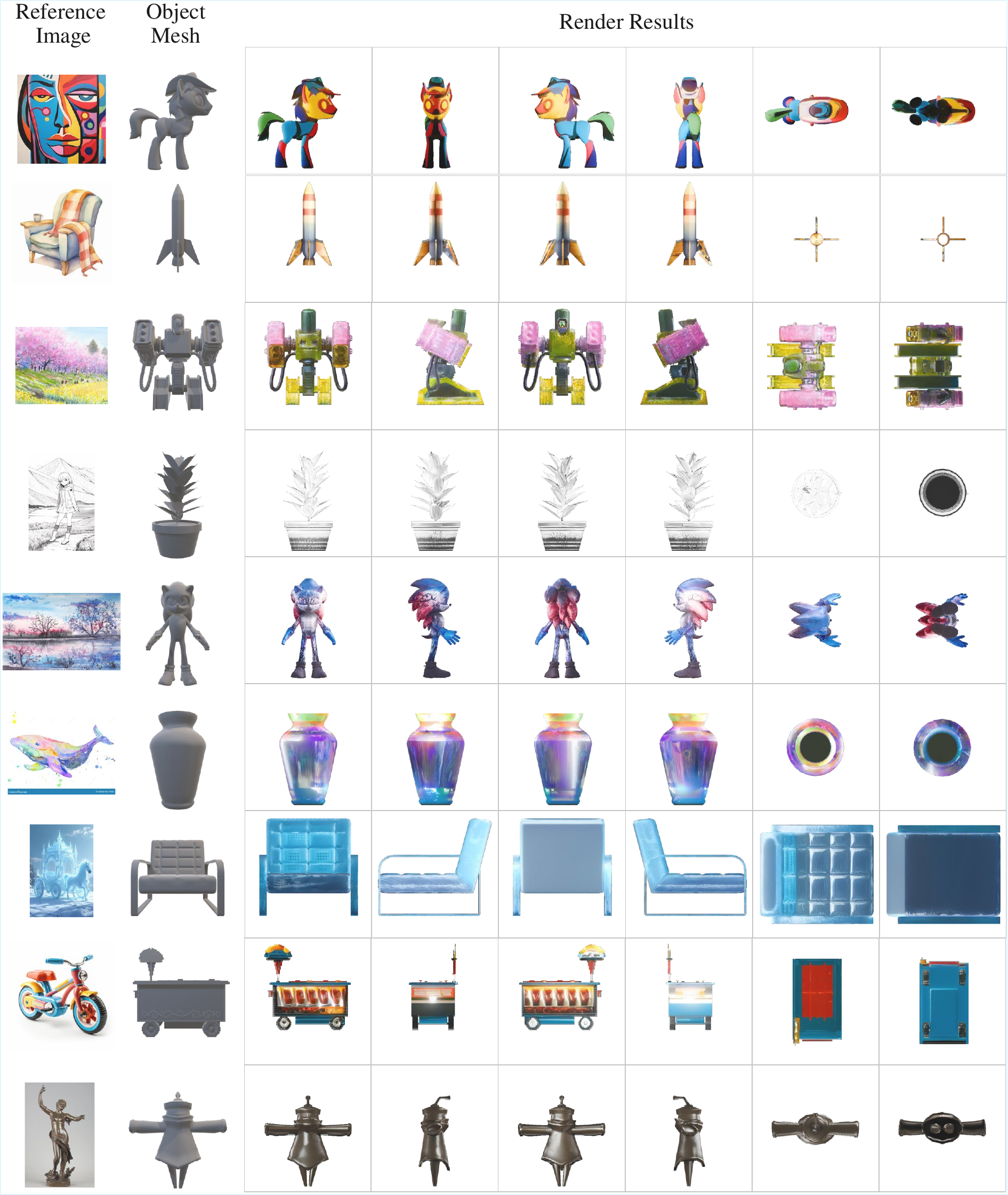}
\caption{
\textbf{More 3D Stylization Results of our method.} Multi-view rendering results by applying various reference styles to a wide range of object meshes from the Objaverse dataset.
}
\label{supp:collected_data_more_result_part_2}
\end{figure}

\begin{figure}[t]
\centering
\includegraphics[width=0.98\textwidth]{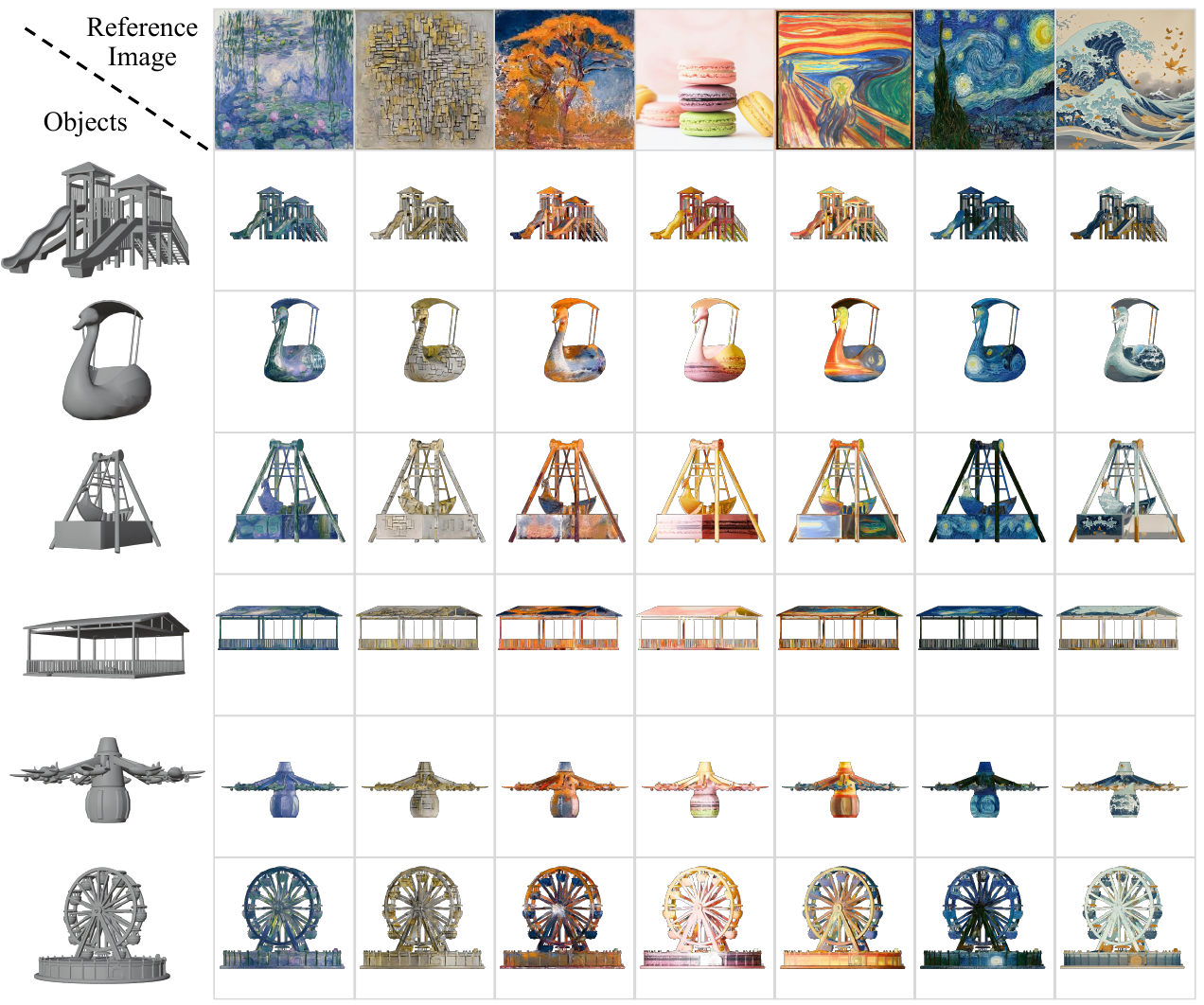}
\caption{
\textbf{Additional qualitative results of our method applied to park scene objects and artistic styles.}
}
\label{supp:artist_painting_more_result}
\end{figure}

\textbf{Further Qualitative Results of Our Methods.} We present additional multi-view rendering results of our method in~\Cref{supp:collected_data_more_result_part_1} and~\Cref{supp:collected_data_more_result_part_2}.~\Cref{supp:collected_data_more_result_part_1} illustrates the outcomes of stylizing a house object using a collection of reference images with distinct artistic styles.~\Cref{supp:collected_data_more_result_part_2} extends this by applying different stylistic references to a diverse set of objects. Both sets of results highlight our method's ability to maintain strong stylistic consistency across multiple viewpoints. Besides,~\Cref{supp:artist_painting_more_result} presents our method's generalization capability by applying celebrated artistic styles to amusement park scene objects. For artistic stylization, we select representative works including Monet's Water Lilies, Piet Mondrian's compositions, Edvard Munch's The Scream, Van Gogh's The Starry Night, and Ukiyo-e prints. The results show stylistic consistency and overall harmony.

\begin{figure}[t]
\centering
\includegraphics[width=0.98\textwidth]{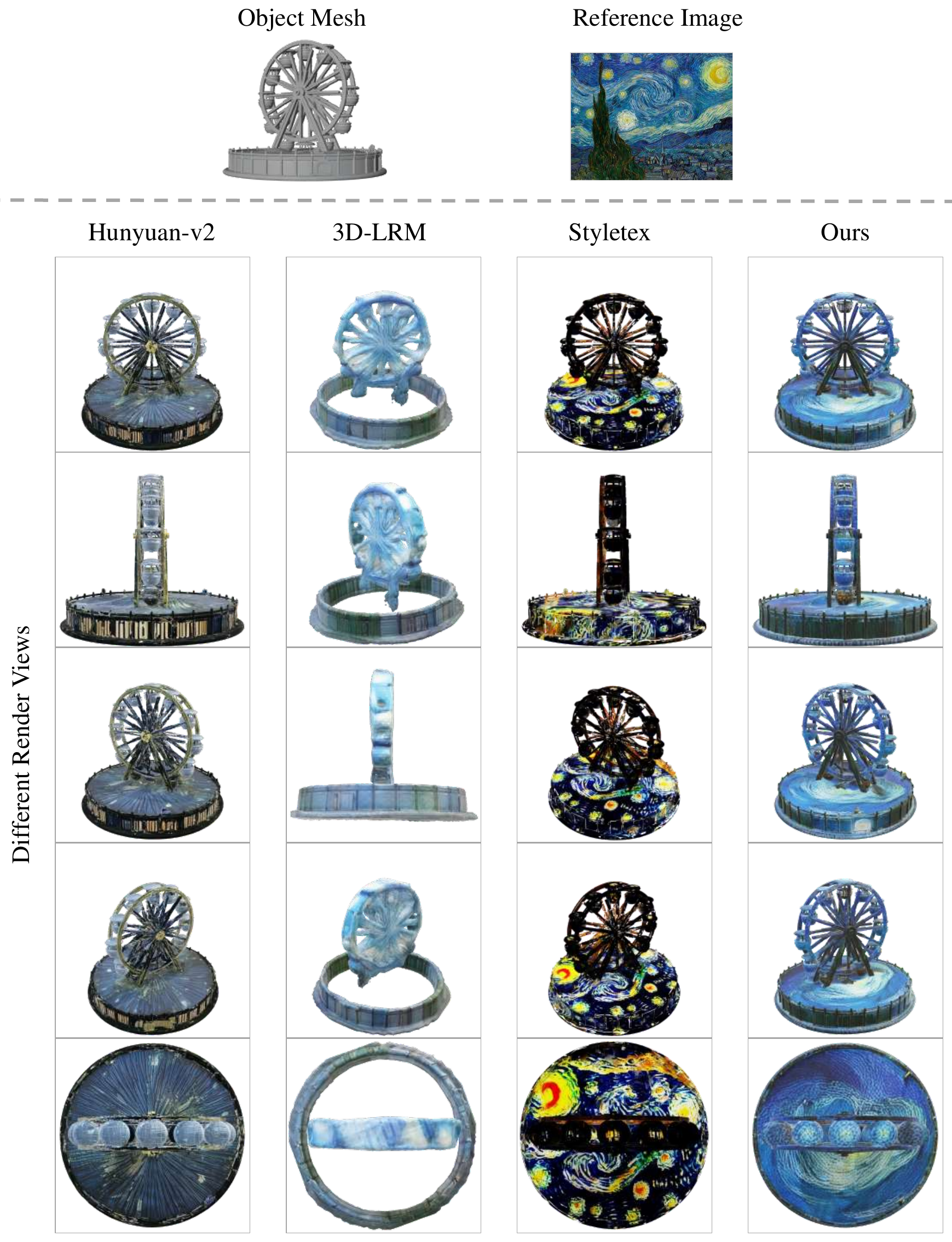}
\caption{
\textbf{Qualitative comparison of multi-view stylization results.} Results below the dashed line show outputs from different methods across multiple viewpoints.
}
\label{supp:supp-more-comparsion-p1}
\end{figure}

\begin{figure}[t]
\centering
\includegraphics[width=0.98\textwidth]{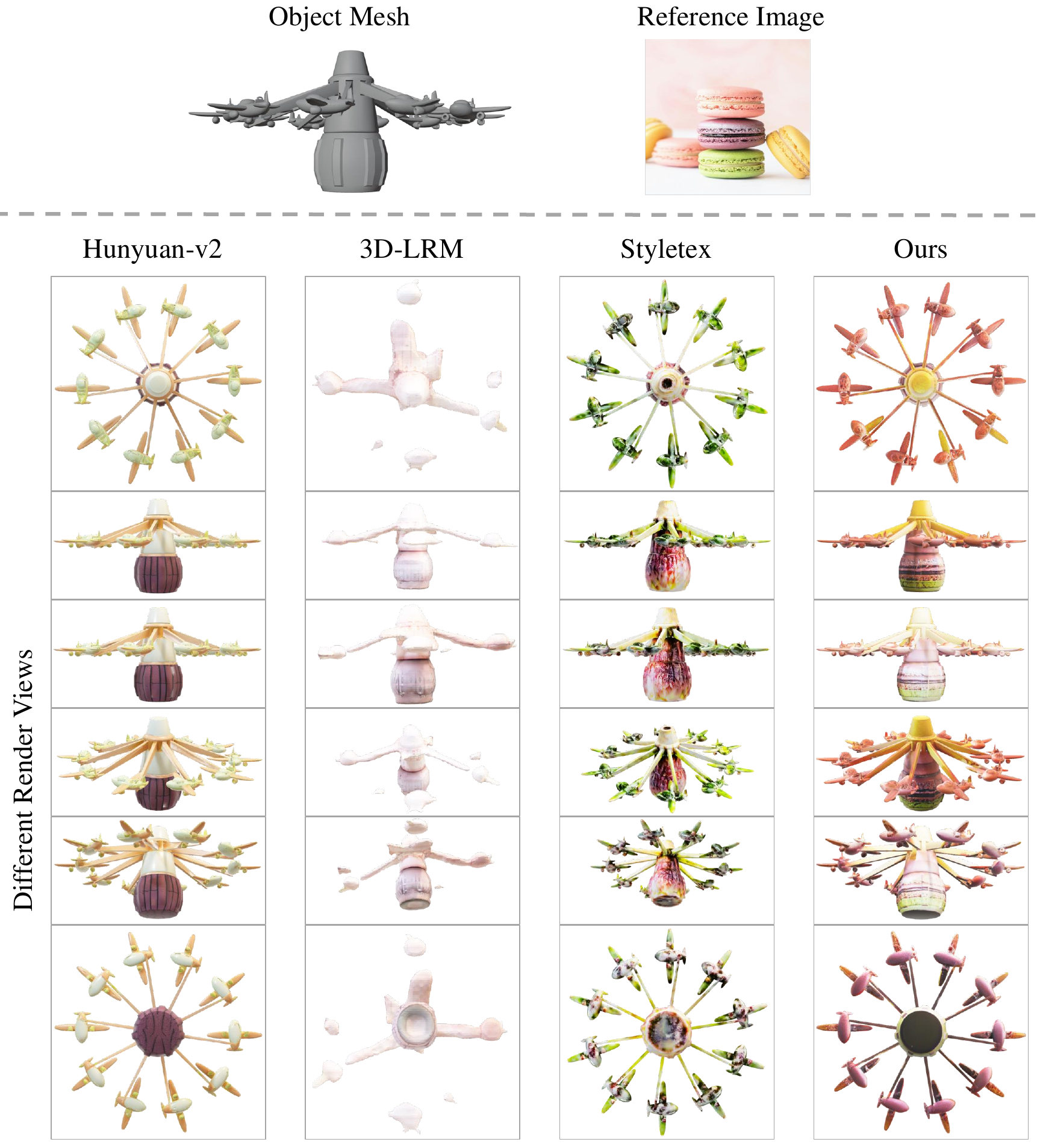}
\caption{
\textbf{Qualitative comparison of multi-view stylization results.} Results below the dashed line show outputs from different methods across multiple viewpoints.
}
\label{supp:supp-more-comparsion-p2}
\end{figure}

\textbf{Further Qualitative Comparisons with SOTA.}~\Cref{supp:supp-more-comparsion-p1} and~\Cref{supp:supp-more-comparsion-p2} present qualitative comparisons with other methods, including Hunyuan-V2~\citep{zhao2025hunyuan3d} as an additional baseline.

\subsection{More Ablation study} 
\label{supp-ablation-study}

\begin{figure}[t]
\centering
\includegraphics[width=0.98\textwidth]{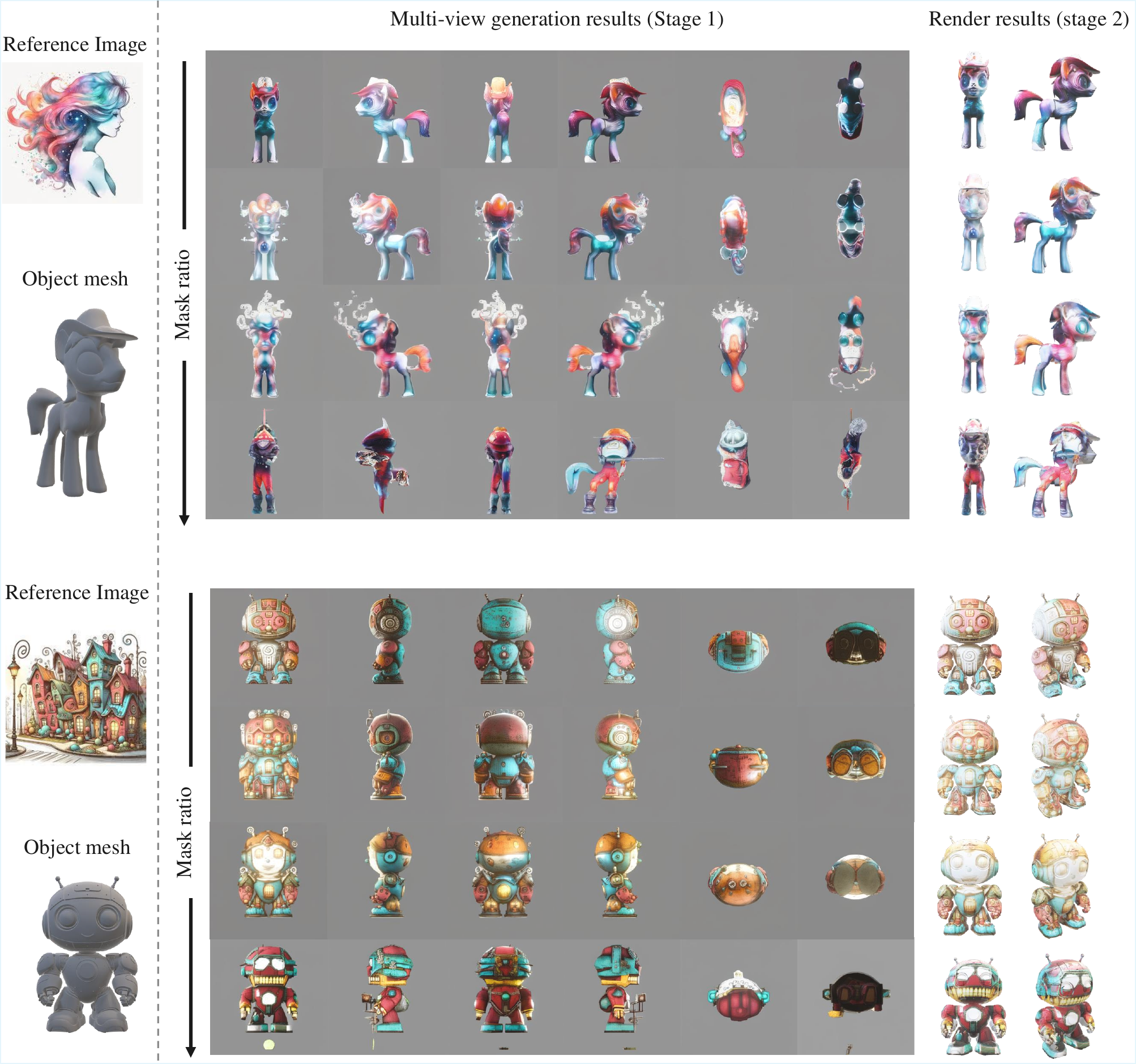}
\caption{
\textbf{Ablation study on the mask ratio $p$.} This figure illustrates the impact of different mask ratio settings on our two-stage process. The panels on the right side of the dashed line show the multi-view generation results and the final rendered results. Each row corresponds $p$ set to a specific ratio: 0.0, 0.25, 0.5, and 0.75, from top to bottom.
}
\label{supp:mask-ratio}
\end{figure}

\textbf{Ablation on Mask Ratio $p$ in~\eqref{eq:jigsaw-2}.} In~\Cref{supp:mask-ratio}, we conduct an ablation study to analyze the impact of the mask ratio. The mask ratio is designed to encourage the model to learn diverse and enhanced feature representations from masked reference images. The results show employing an appropriate mask ratio can enhance the diversity of the learned feature expressions, while setting it too high (e.g., 0.75) proves detrimental. 
An excessive mask ratio leads to lost geometric information with generated multi-views, which in turn causes an uneven appearance with artifacts during the baking process.

\begin{figure}[t]
    \centering
    \includegraphics[width=0.85\textwidth]{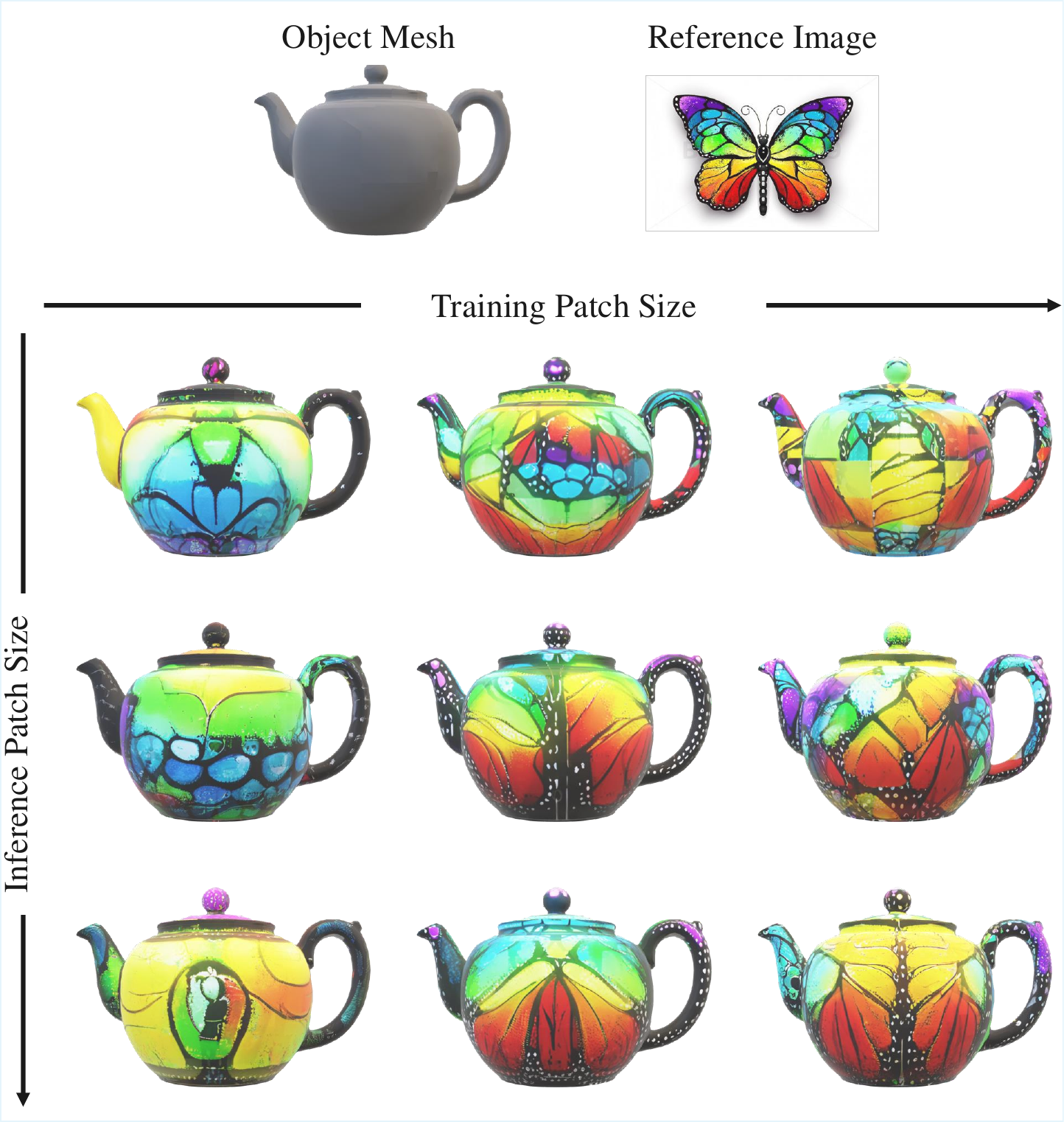} 
    \caption{
        \textbf{Ablation study on patch size $S$.} We analyze the impact of varying patch sizes during training and inference. The entire image size is fixed at $512 \times 512$, and the image patch size is set to $S \times S$. From left to right, the columns show rendering results using training patch sizes of $32 \times 32$, $64 \times 64$, and $128 \times 128$, respectively. From top to bottom, the rows correspond to inference patch sizes of $32 \times 32$, $64 \times 64$, $128 \times 128$.}
    \label{fig:patch_ablation}
\end{figure}

\textbf{Ablation on Training and Inference Patch Size $S$ in~\eqref{eq:jigsaw-1}.} The image patch serves as the primary carrier of style information, making its size a critical hyperparameter.~\Cref{fig:patch_ablation} showcases the qualitative rendering results of our model under different patch size configurations. A training patch size that is too small may not sufficiently contain the style information, whereas a size that is too large can cause the model to focus excessively on content details. We find that for a majority of images, a training patch size of $64 \times 64$ coupled with an inference patch size of $128 \times 128$ produces the stable results. This is likely due to the nature of the data domain, where this configuration strikes an optimal balance.

\subsection{More Application results} 
\label{supp-more-application}
\begin{figure}[t]
\centering
\includegraphics[width=0.98\textwidth]{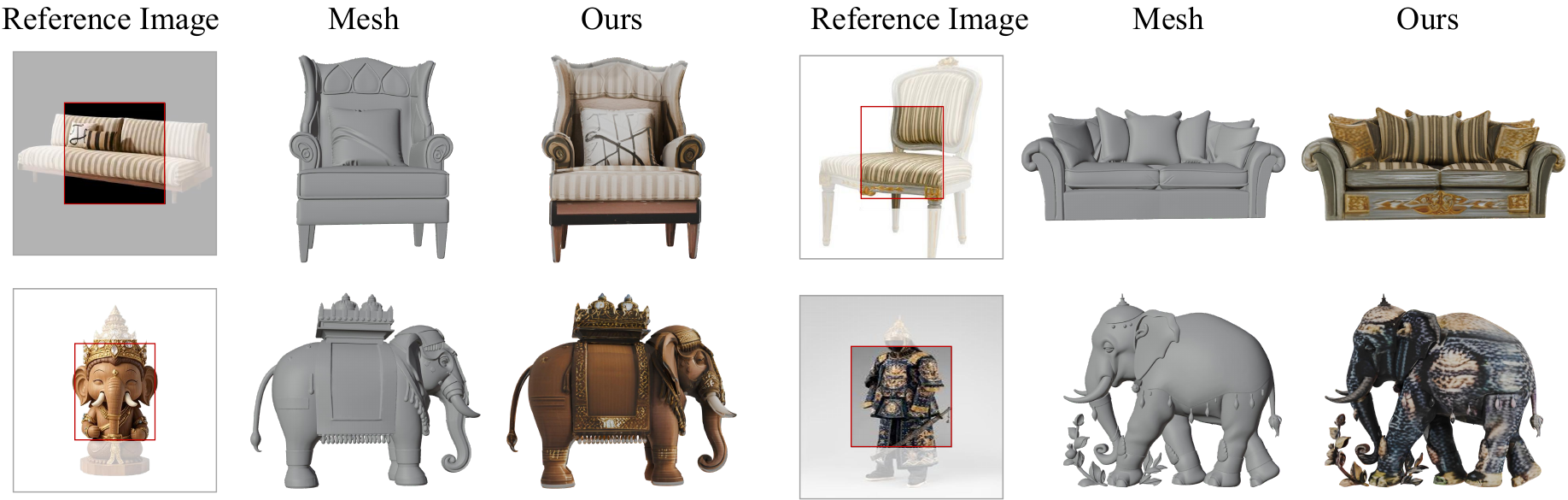}
\caption{
\textbf{Partial stylization results.} For reference image, we preserve the partial region (as presented as red box) and mask other regions for partial inference.
}
\label{fig:more-application-3}
\end{figure}

\textbf{Partial Object Stylization.} To evaluate our method's ability to understand and transfer style attributes from limited visual cues, we conduct experiments using only cropped regions of the reference image. As shown in \Cref{fig:more-application-3}, our approach can successfully infer a globally consistent style from a partial reference and apply it coherently to the full 3D object. This demonstrates the robustness of our model in handling partial style references while maintaining semantic and stylistic coherence.

\begin{figure}[t]
\centering
\includegraphics[width=0.98\textwidth]{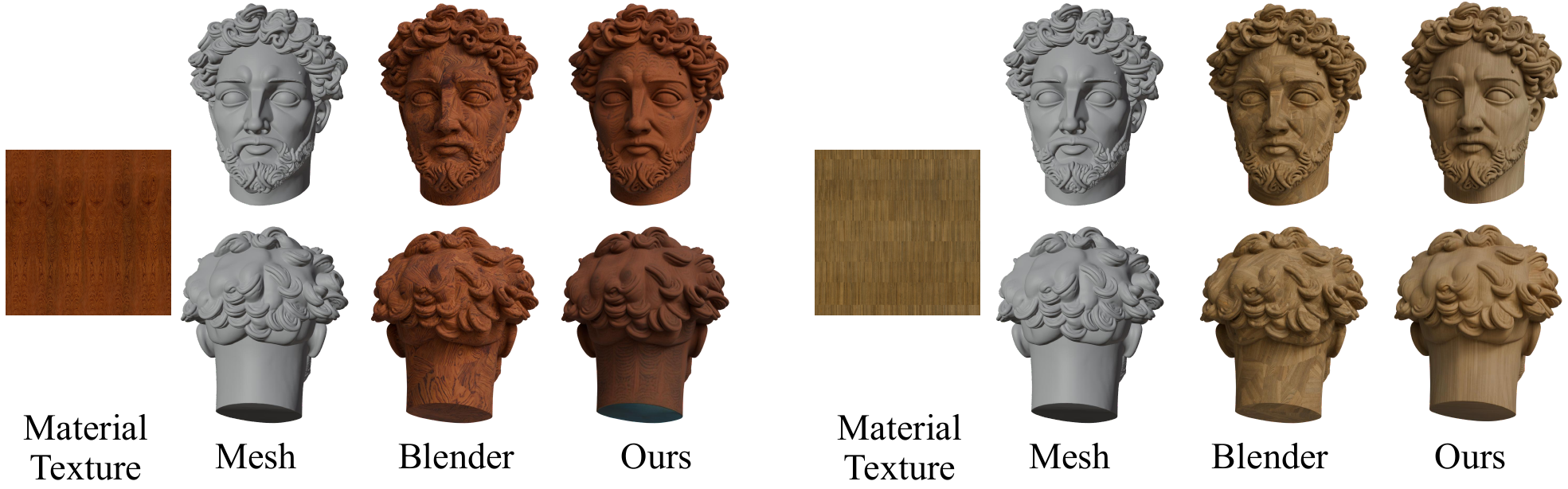}
\caption{
\textbf{Comparison with conventional renderers on tileable texture generation.}
Our method effectively eliminates seam artifacts compared to Blender, which often produces inconsistent texture directionality across UV boundaries.
}
\label{fig:more-application-2}
\end{figure}

\textbf{Tileable Texture Generation.} 
We further demonstrate that our method is not capable of transferring styles but also can create consistent texture tiles from an example image.
%
As shown in~\Cref{fig:more-application-2}, in comparison with the direct texture mapping in Blender with noticeable irregularities and artifacts, especially around UV seams, our jigsaw-based strategy disrupts spatial biases while maintaining strong consistency and continuity across the entire 3D surface.

\end{document}